\documentclass{article}

\usepackage[utf8]{inputenc}
\usepackage{microtype}
\usepackage{graphicx}
\usepackage{subcaption}
\usepackage{booktabs} 
\usepackage{rotating}
\usepackage{hyperref}

\usepackage{tikz}
\usepackage{pgfplots}

\usetikzlibrary{calc}
\usetikzlibrary{positioning}
\usepgflibrary{fpu}
\usepgfplotslibrary{fillbetween}

\pgfplotsset{compat=newest}

\usepackage{placeins}

\newenvironment{customlegend}[1][]{%
    \begingroup
    \csname pgfplots@init@cleared@structures\endcsname
    \pgfplotsset{#1}%
}{%
    \csname pgfplots@createlegend\endcsname
    \endgroup
}%
\def\addlegendimage{\csname pgfplots@addlegendimage\endcsname}

\usepackage[accepted]{icml2020}

\icmltitlerunning{Evaluating Uncertainty Estimation Methods on 3D Semantic Segmentation of Point Clouds}

\usepackage{placeins}

\definecolor{unlabeled}{RGB}{0, 0, 0}
\definecolor{car}{RGB}{100, 150, 245}
\definecolor{bicycle}{RGB}{100, 230, 245}
\definecolor{motorcycle}{RGB}{30, 60, 150}
\definecolor{truck}{RGB}{80, 30, 180}
\definecolor{othervehicle}{RGB}{100, 80, 250}
\definecolor{person}{RGB}{255, 30, 30}
\definecolor{bicyclist}{RGB}{255, 40, 200}
\definecolor{motorcyclist}{RGB}{150, 30, 90}
\definecolor{road}{RGB}{255, 0, 255}
\definecolor{parking}{RGB}{231, 17, 35}
\definecolor{sidewalk}{RGB}{255, 0, 0}
\definecolor{otherground}{RGB}{175, 0, 75}
\definecolor{building}{RGB}{255, 200, 0}
\definecolor{fence}{RGB}{255, 120, 50}
\definecolor{vegetation}{RGB}{0, 175, 0}
\definecolor{trunk}{RGB}{135, 60, 0}
\definecolor{terrain}{RGB}{150, 240, 80}
\definecolor{pole}{RGB}{150, 240, 255}
\definecolor{trafficsign}{RGB}{255, 0, 0}

\definecolor{entropyA}{RGB}{0, 255, 0}
\definecolor{entropyB}{RGB}{186, 216, 104}
\definecolor{entropyC}{RGB}{0, 178, 148}
\definecolor{entropyD}{RGB}{0, 187, 241}
\definecolor{entropyE}{RGB}{0, 24, 143}
\definecolor{entropyF}{RGB}{104, 33, 121}
\definecolor{entropyG}{RGB}{236, 0, 140}
\definecolor{entropyH}{RGB}{255, 241, 0}
\definecolor{entropyI}{RGB}{255, 140, 0}
\definecolor{entropyJ}{RGB}{255, 0, 0}

\begin{document}

\twocolumn[
\icmltitle{Evaluating Uncertainty Estimation Methods on 3D Semantic Segmentation of Point Clouds}



\icmlsetsymbol{equal}{*}

\begin{icmlauthorlist}
\icmlauthor{Swaroop Bhandary K}{hbrs}
\icmlauthor{Nico Hochgeschwender}{hbrs}
\icmlauthor{Paul Plöger}{hbrs}
\icmlauthor{Frank Kirchner}{dfki}
\icmlauthor{Matias Valdenegro-Toro}{dfki}
\end{icmlauthorlist}

\icmlaffiliation{hbrs}{Department of Computer Science, Hochschule Bonn Rhein Sieg, Bonn, Germany}
\icmlaffiliation{dfki}{German Research Center for Artificial Intelligence, Bremen, Germany}

\icmlcorrespondingauthor{Swaroop Bhandary K}{swaroop1904@gmail.com}
\icmlcorrespondingauthor{Matias Valdenegro-Toro}{matias.valdenegro@dfki.de}

\icmlkeywords{3D semantic Segmentation, Bayesian Deep Learning, Uncertainty Estimation}

\vskip 0.3in
]



\printAffiliationsAndNotice{}  

\begin{abstract}
	
Deep learning models are extensively used in various safety critical applications. Hence these models along with being accurate need to be highly reliable. One way of achieving this is by quantifying uncertainty. Bayesian methods for UQ have been extensively studied for Deep Learning models applied on images but have been less explored for 3D modalities such as point clouds often used for Robots and Autonomous Systems. In this work, we evaluate three uncertainty quantification methods namely Deep Ensembles, MC-Dropout and MC-DropConnect on the DarkNet21Seg 3D semantic segmentation model and comprehensively analyze the impact of various parameters such as number of models in ensembles or forward passes, and drop probability values, on task performance and uncertainty estimate quality. We find that Deep Ensembles outperforms other methods in both performance and uncertainty metrics. Deep ensembles outperform other methods by a margin of 2.4\% in terms of mIOU, 1.3\% in terms of accuracy, while providing reliable uncertainty for decision making.
\end{abstract}

\section{Introduction}

Deep learning based models are being used extensively for several computer vision tasks such as object classification, detection and segmentation. Despite their popularity, most of these models do not consider the uncertainty at their output, producing overconfident probabilities for incorrect predictions, and generally are not being able to predict their accuracy, specially for out of distribution data.

Additional measures are required to make models more reliable. Bayesian inference provides a solution to this issue by predicting a distribution rather than making point estimates. However, these models are computationally expensive. Various non-bayesian methods such as ensembling \cite{lakshminarayanan2017simple} or stochastic regularization techniques like Dropout \cite{gal2016dropout},  DropConnect \cite{mobiny2019dropconnect}, and Stochastic Batch Normalization \cite{atanov2019uncertainty}  have been used to obtain the uncertainty estimate from machine learning models. 


For applications such as self driving cars and autonomous robotics, a 3D understanding of the environment is necessary. For such tasks, 3D sensors such as LiDARs are used since the data captured by these sensor contain depth information which is an important feature for these tasks and can capture the geometric properties of the objects and are more robust to lighting conditions. However dealing with 3D data comes with its own set of challenges. Point clouds, one of the most common representations of 3D data, is an unstructured data type. Hence traditional CNN models which have shown to perform extremely well for images in various computer vision tasks due to the well structured grid like patterns cannot be directly applied to point clouds \cite{qi2017pointnet}. 

Point clouds need to be processed to extract valuable information regarding the environment. One way of extracting this for a given scene is semantic segmentation. This involves point-wise classification given a set of classes. Given a semantically segmented point cloud, a variety of information can be extracted from it. For example, the objects present in the scene and also a rough estimate of the position of each object present in the scene. In autonomous driving, semantic segmentation is used to estimate the drivable space and also determine lane boundaries.


To the best of our knowledge, most methods for semantic segmentation of 3D point clouds do not consider uncertainty, which is particularly important for the safe use of learned models, in particular regarding detecting novel situations that are not considered in the training sets of the model. There are various instances where there have been serious repercussions of using deep learning models in real world scenarios. Two of the well known cases in autonomous driving include Tesla's autopilot failure, where the an autonomous vehicle misclassified the white carrier with clear sky and crashed into it \cite{tesla} and Uber's failure to detect pedestrian crossing the road which resulted in the car hitting the pedestrian leading to a fatality \cite{uber}.

In this work, we select one state of the art 3D segmentation model based on their performance in the SemanticKITTI dataset, namely DarkNet21Seg \cite{behley2019semantickitti}, and perform a systematic evaluation of three uncertainty based approaches which work as an approximate Bayesian Neural Network on the model. We find that Deep Ensembles is particularly well suited for this problem. 

\section{Related work}

\subsection{3D semantic segmentation}

Some approaches for processing 3D data involve pre-processing them to convert the point clouds into a structured representation either by projecting into a 2D plane creating a range image \cite{wu2019squeezesegv2}; \cite{milioto2019rangenet++} or by voxelizing them \cite{tchapmi2017segcloud}. 2D and 3D convolutional models are then applied respectively on top of these structured pattern.

A few other methods treat point clouds as graphs and apply graph neural networks \cite{qi20173d}.

One of the earliest architectures that directly process point clouds is PointNet \cite{qi2017pointnet} and its extension PointNet++ \cite{qi2017pointnet++} \cite{engelmann2017exploring}.

\subsection{Uncertainty estimation}
Various approaches exist to produce uncertainty estimates from neural network models. They include applying stochastic regularization techniques (SRT) during test time as well such as MC-Droput \cite{gal2016dropout}; MC-DropConnect \cite{mobiny2019dropconnect}; and Stochastic Batch Normalization \cite{atanov2019uncertainty}. 

It was shown that a model trained with SRT such as DropConnect and Dropout is an approximation of a Deep Gaussian Process. Hence by turning on the SRT during test time, we get an uncertainty value that is an approximation to the posterior distribution without any change to the network or training procedure.


\begin{figure}[t!]
    \centering
    \includegraphics[scale=0.3]{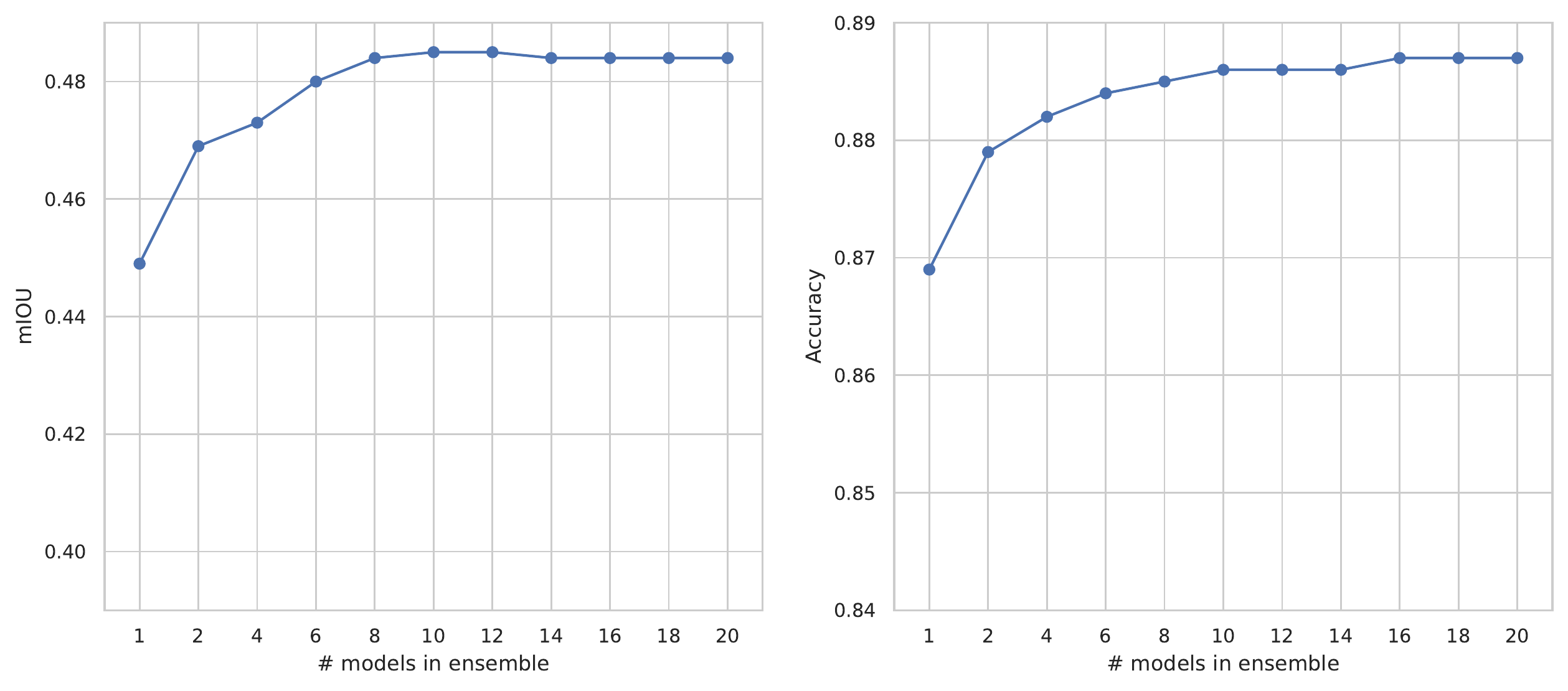}
    \caption{Task Performance vs \# of Models in Ensemble}
    \label{figure:ens_per}
\end{figure}
\begin{figure}[t!]
    \centering
    \includegraphics[scale=0.30]{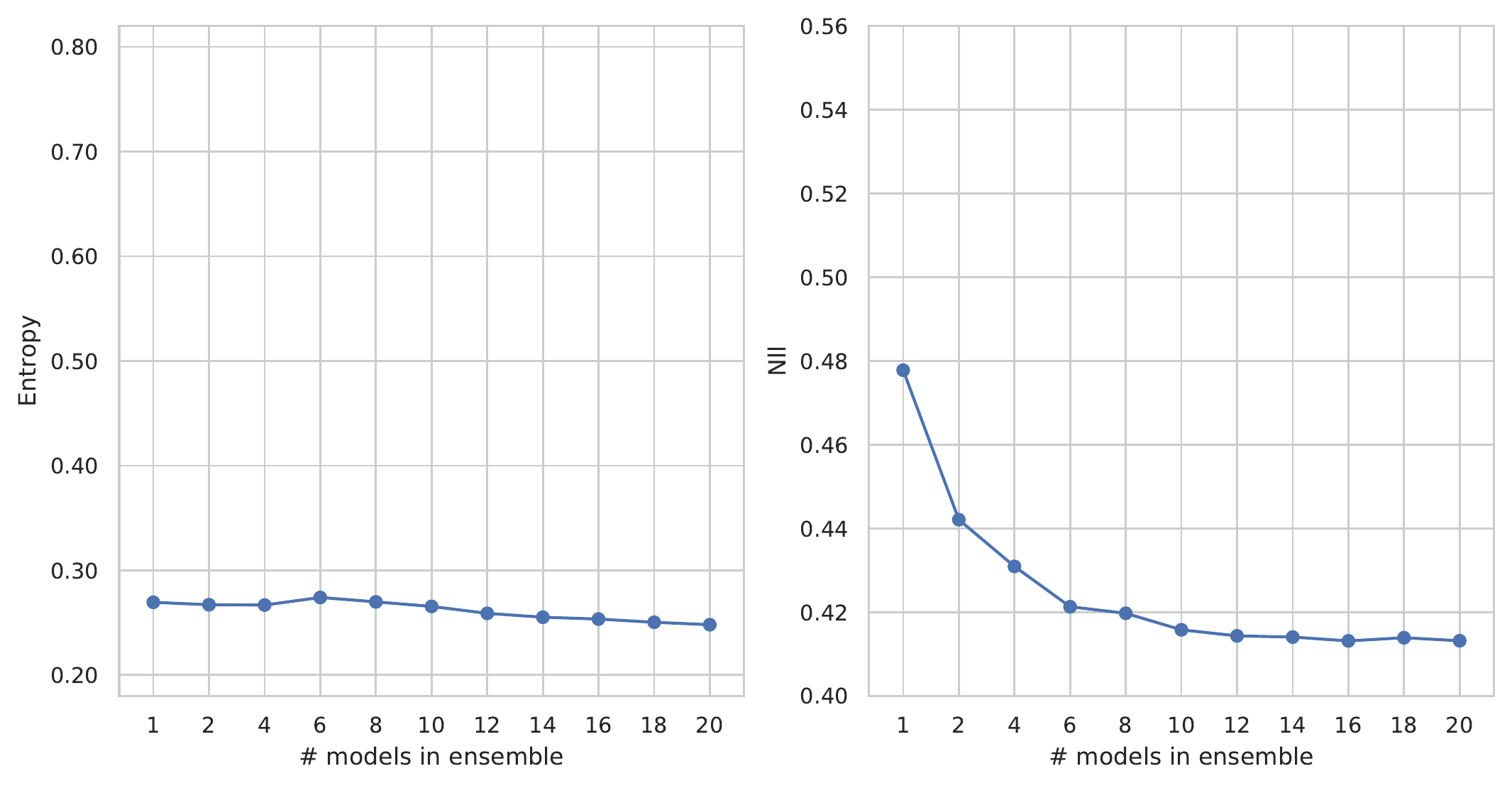}
    \caption{Uncertainty vs \# of Models in Ensemble}
    \label{figure:ens_unc}
\end{figure}

\section{Proposed Methodology}

We perform a thorough evaluation of the uncertainty estimation methods on the DarkNet21Seg architecture \cite{behley2019semantickitti} due to its state of the art performance on the SemanticKITTI dataset. It is a U-net based architecture with five residual encoders and five decoder blocks. The point cloud is transformed into an image by performing a spherical projection, in a similar way as SqueezeSeg \cite{wu2019squeezesegv2}.

Weighted cross entropy loss function has been used during training as the SemanticKITTI dataset is unbalanced. The learning rate used is $10^{-3}$ with a decay of 0.99 after every epoch. 

As the labels for the test sequence is not publicly available, we use sequence 08 which is the recommended validation set and consists of 4071 point clouds each consisting of around 100K points as the test sequence. 

We apply three uncertainty based methods namely Deep Ensembles, MC Dropout, and MC DropConnect and make comparisons based on the following metrics: mIOU and pixel accuracy to measure performance of the network on the semantic segmentation task; entropy, negative log-likelihood and area under the accuracy vs. confidence curve to measure the uncertainty reliability; and calibration plots to measure how calibrated the network is.

To compare various methods we average the uncertainty values over the validation set for the dataset and also provide class-wise values. 

\begin{figure}[t!]
    \begin{minipage}[t]{0.2\textwidth}	
        \centering
        \includegraphics[scale=0.3]{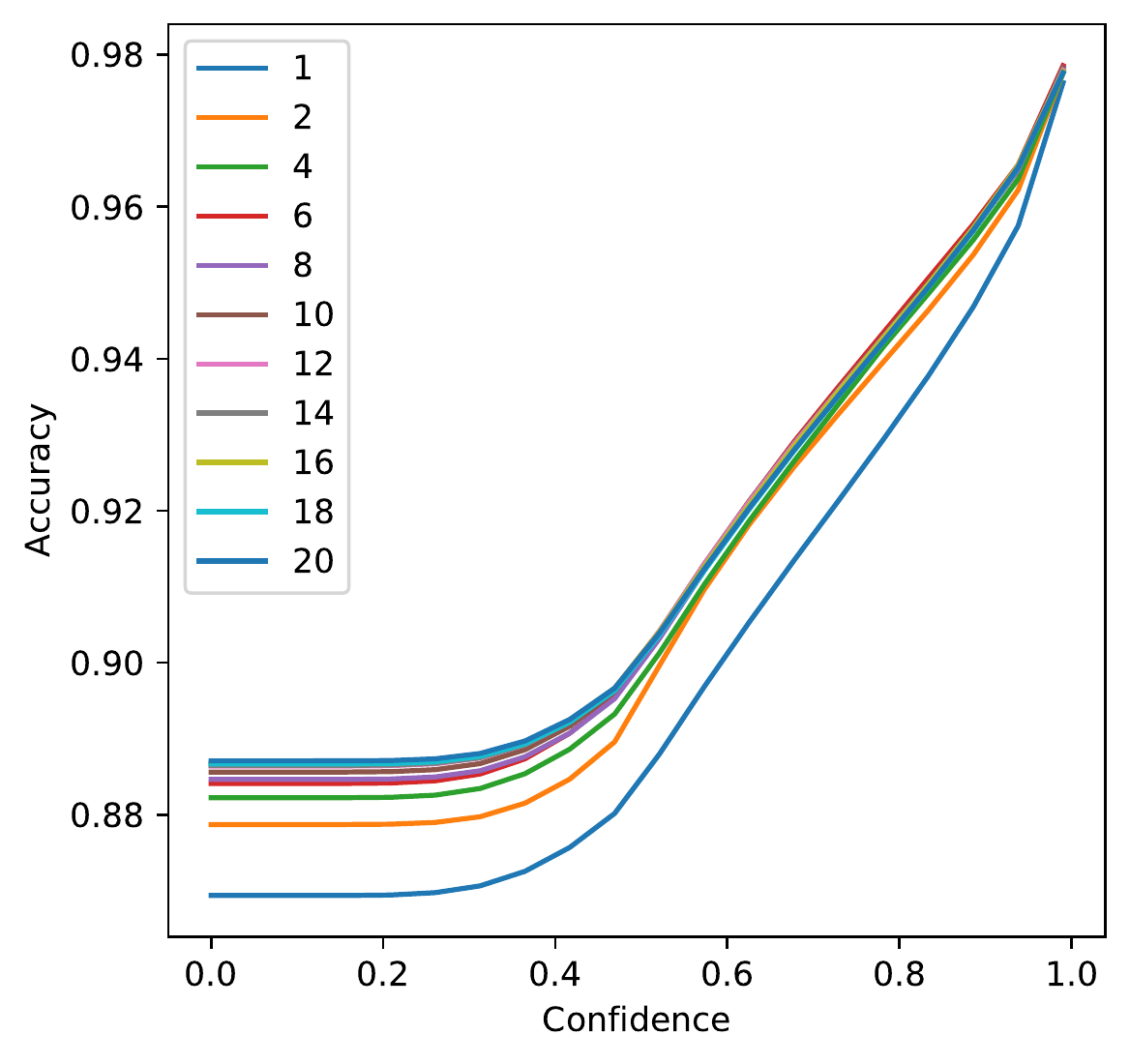}
        \caption{Deep Ensembles - Accuracy vs Confidence}
        \label{figure:ens_acp}
    \end{minipage} \hfill
    \begin{minipage}[t]{0.2\textwidth}	
        \centering
        \includegraphics[scale=0.3]{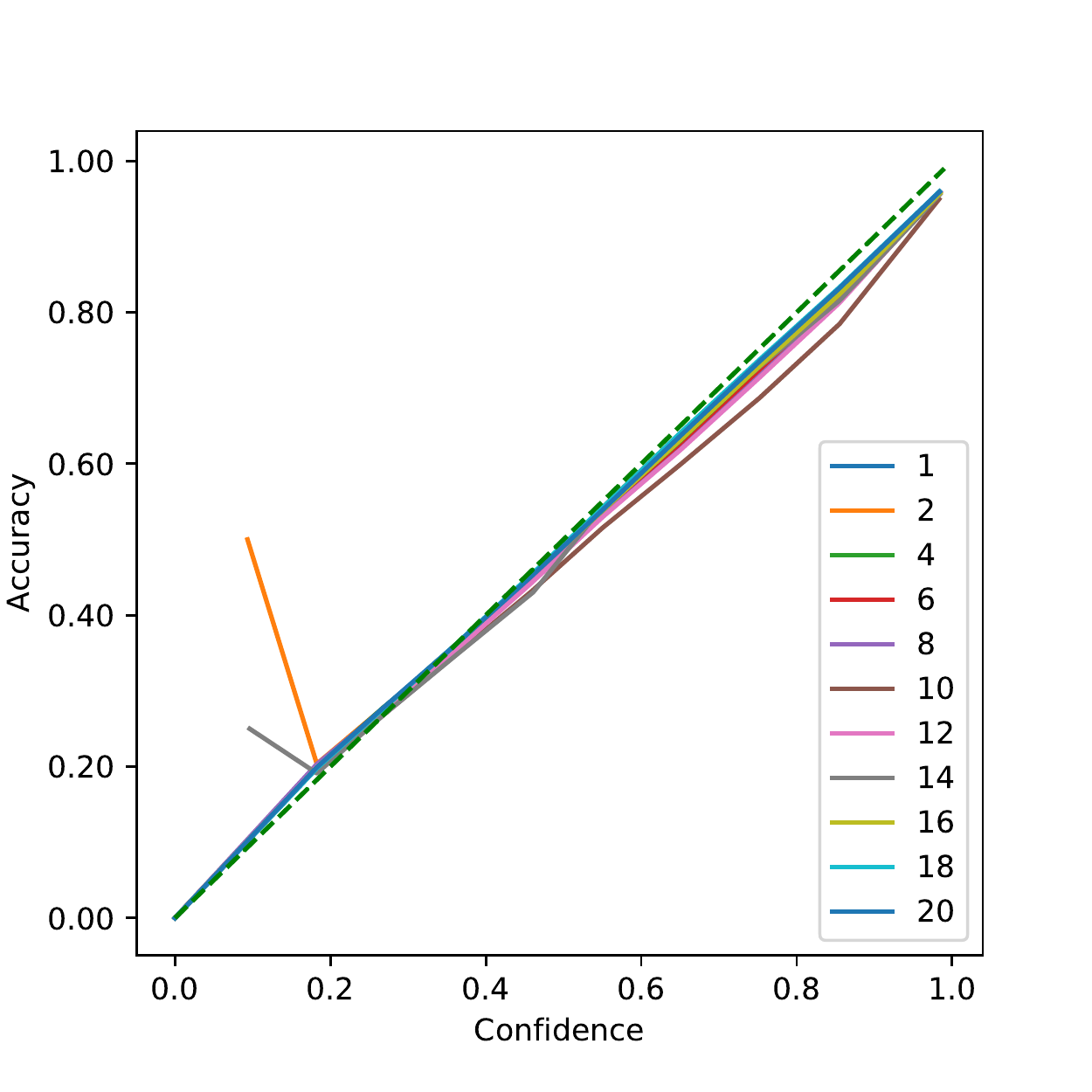}
        \caption{Deep Ensembles - Reliability Plot}
        \label{figure:ens_calib}
    \end{minipage}	
\end{figure}

\begin{figure}[t!]
    \centering
    \includegraphics[scale=0.3]{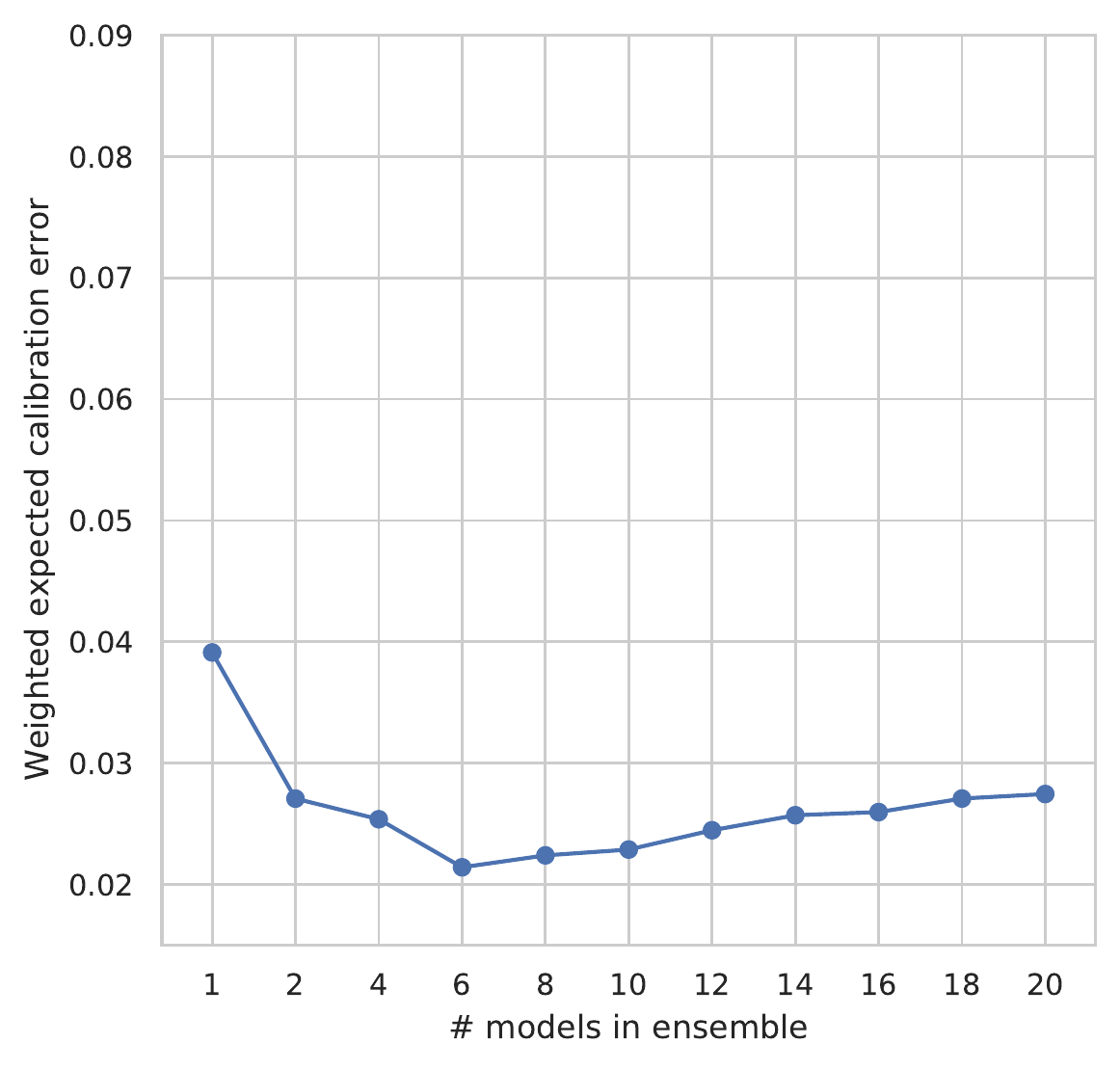}
    \caption{Deep Ensembles - Calibration metrics vs \# Models in Ensemble}
    \label{figure:ens_ece}
\end{figure}

\section{Experimental Evaluation}

\subsection{Deep Ensembles}

The number of models in an ensemble is varied up to 20 in increments of 2, with results shown in Figure \ref{figure:ens_per}. Ensembling is known to boost model task performance and similar trend can be seen for the DarkNet21Seg model as well. All metrics improve with the increase in the number of models in the ensembles. For all the metrics, there are diminishing returns for more than 8 models in the ensemble. 

The area under accuracy vs confidence curve also increases indicating the model is more reliable as ensemble members are added. Comparing the class-wise mIOU and accuracy in table \ref{table:darknet_perf_iou} and \ref{table:darknet_perf_acc} and the class-wise uncertainty in figure \ref{figure:ens_ep_class} and \ref{figure:ens_nll_class} we can see that entropy is lower for classes with higher IOU such as road, car and entropy is higher for classes with lower IOU such as motorcyclist and other-ground.

Calibration captures how reliable the probability predicted by the network are. As seen in figure \ref{figure:ens_calib} and \ref{figure:ens_ece}, the ensemble with more models is better calibrated.

For ensembles the entropy decreases as we increase the number of models, but for MC Dropout when increasing the number of forward passes, the entropy increases. This could be due to the increased stochasticity in Dropout compared to an ensemble.

\begin{figure}[t!]
    \centering
    \includegraphics[scale=0.35]{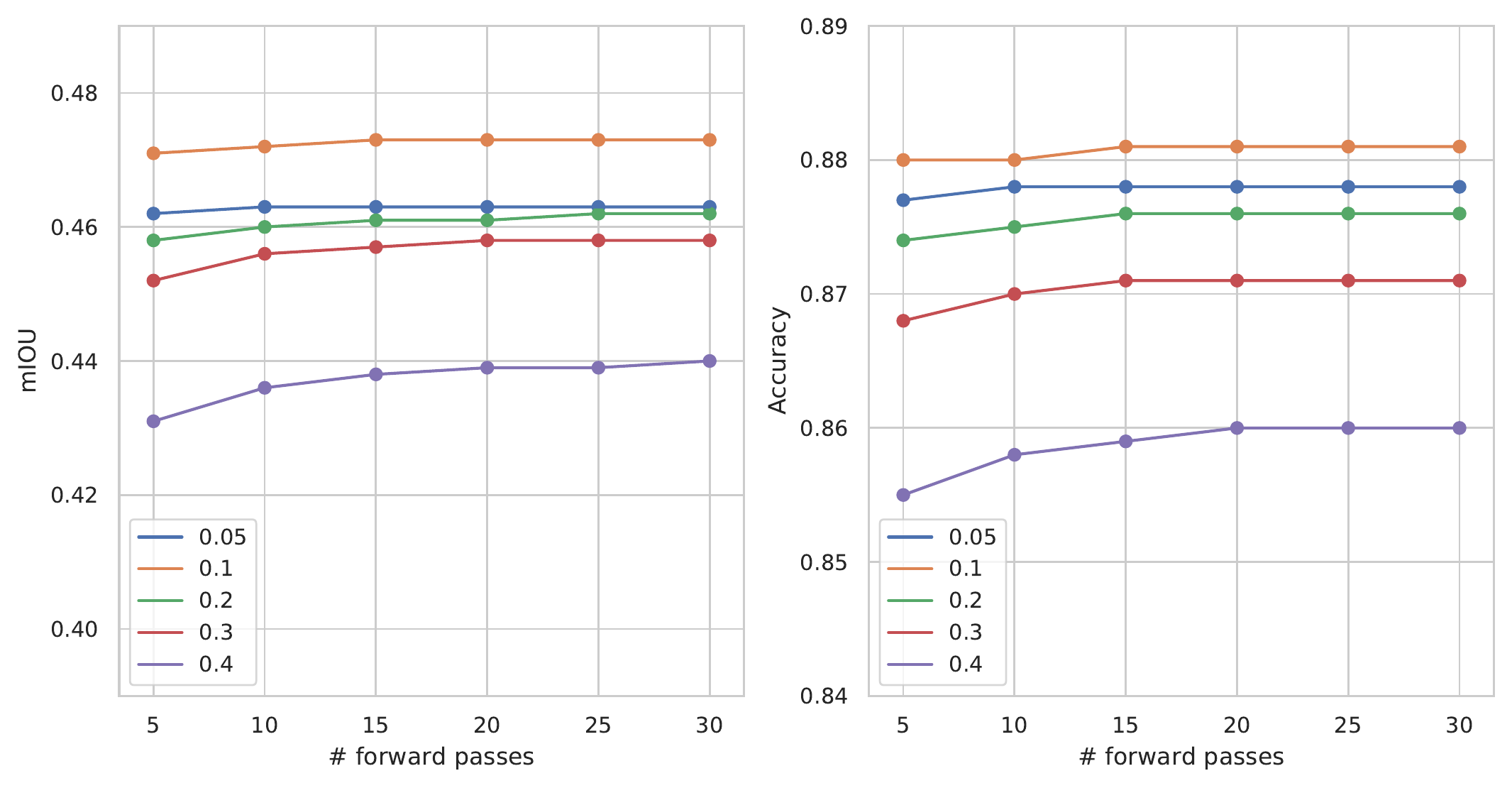}
    \caption{MC-Dropout - Task Performance vs \# of Forward Passes}
    \label{figure:dp_per}
\end{figure}

\begin{figure}[t!]
    \centering
    \includegraphics[scale=0.35]{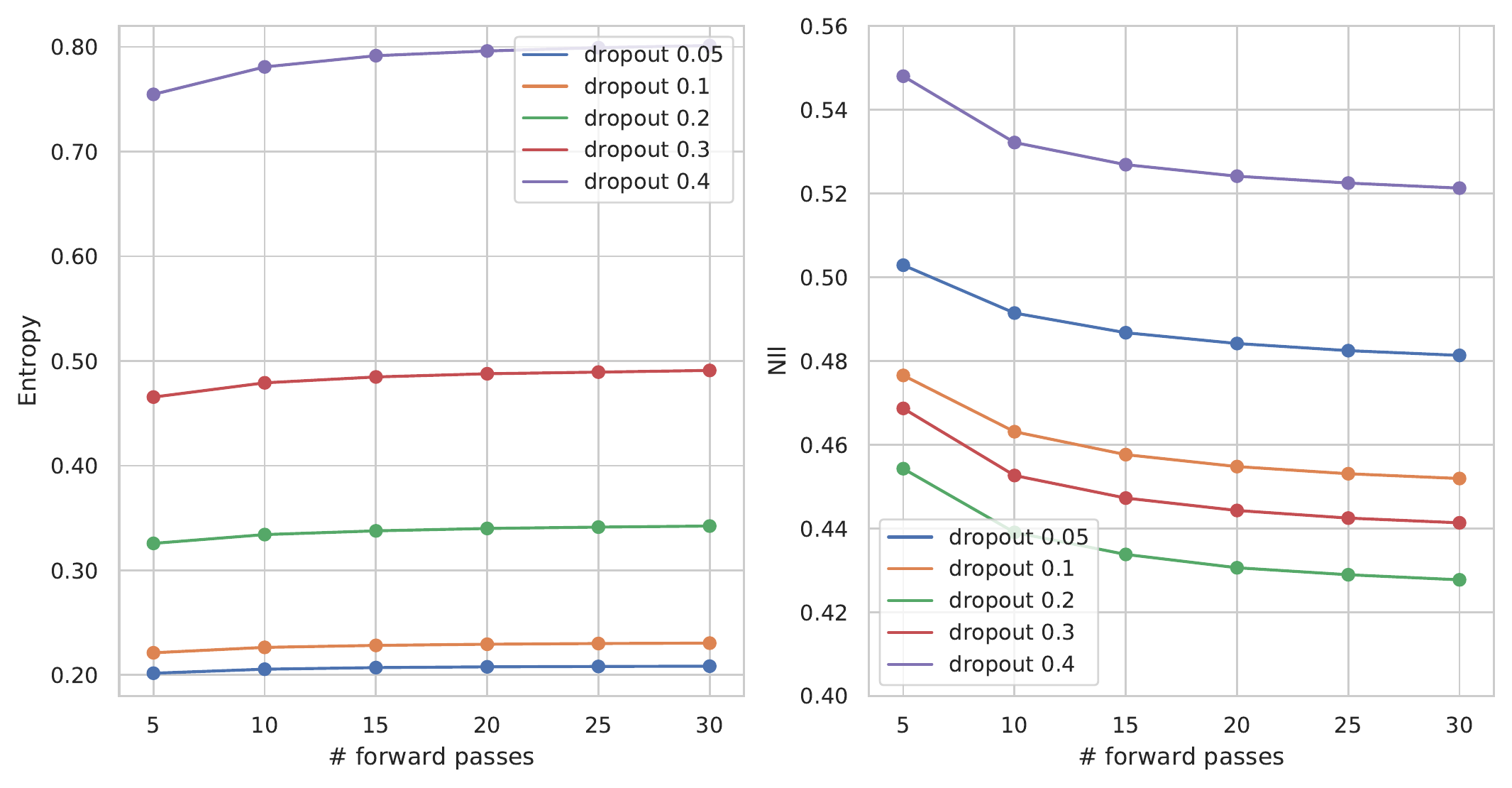}
    \caption{MC-Dropout - Uncertainty vs \# of Forward Passes}
    \label{figure:dp_unc}
\end{figure}

\begin{figure}[t!]
    \begin{minipage}[t]{0.2\textwidth}	
        \centering
        \includegraphics[scale=0.3]{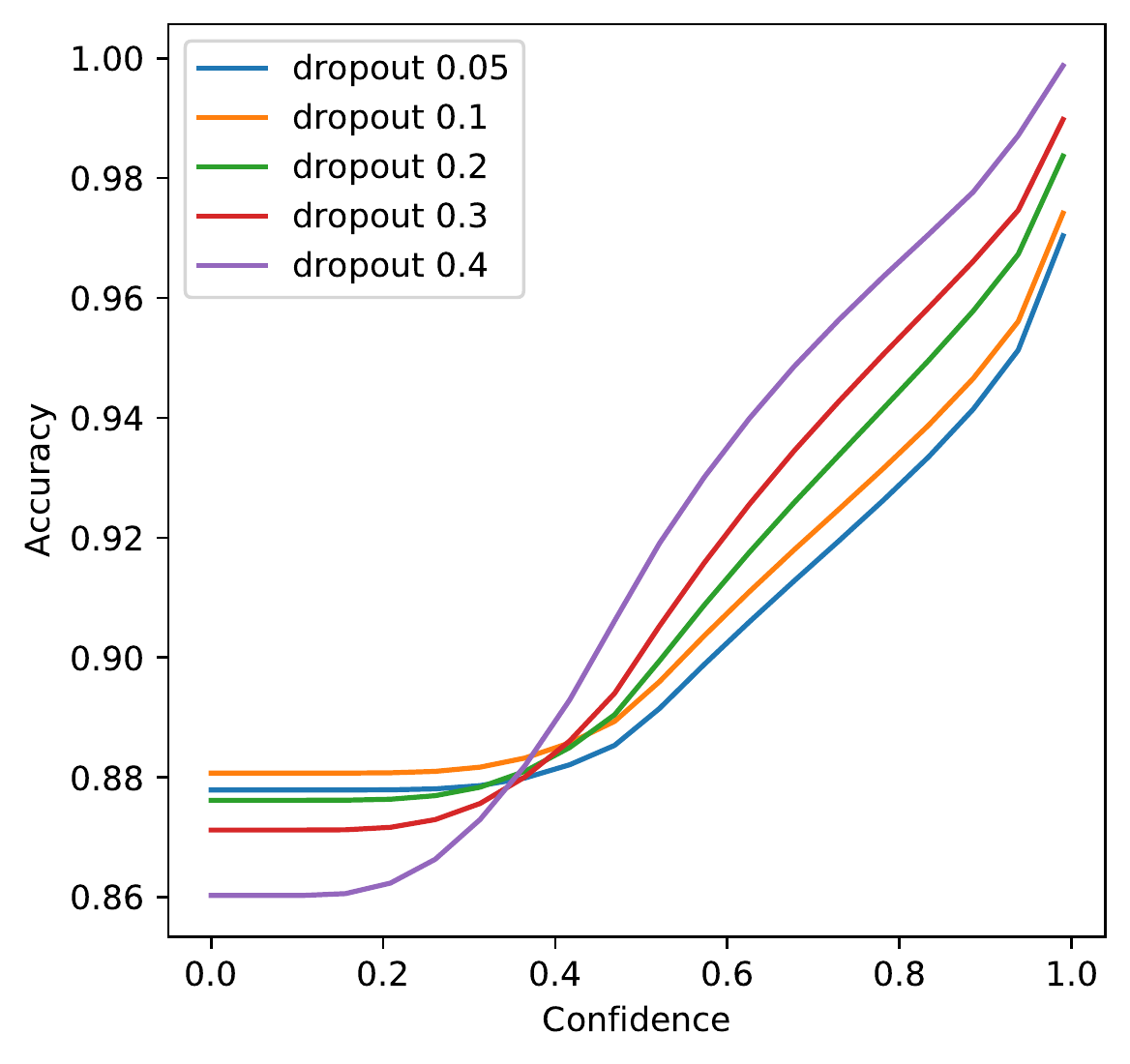}
        \caption{MC-Dropout - Accuracy vs Confidence}
        \label{figure:dp_acp}
    \end{minipage} \hfill
    \begin{minipage}[t]{0.2\textwidth}	
        \centering
        \includegraphics[scale=0.3]{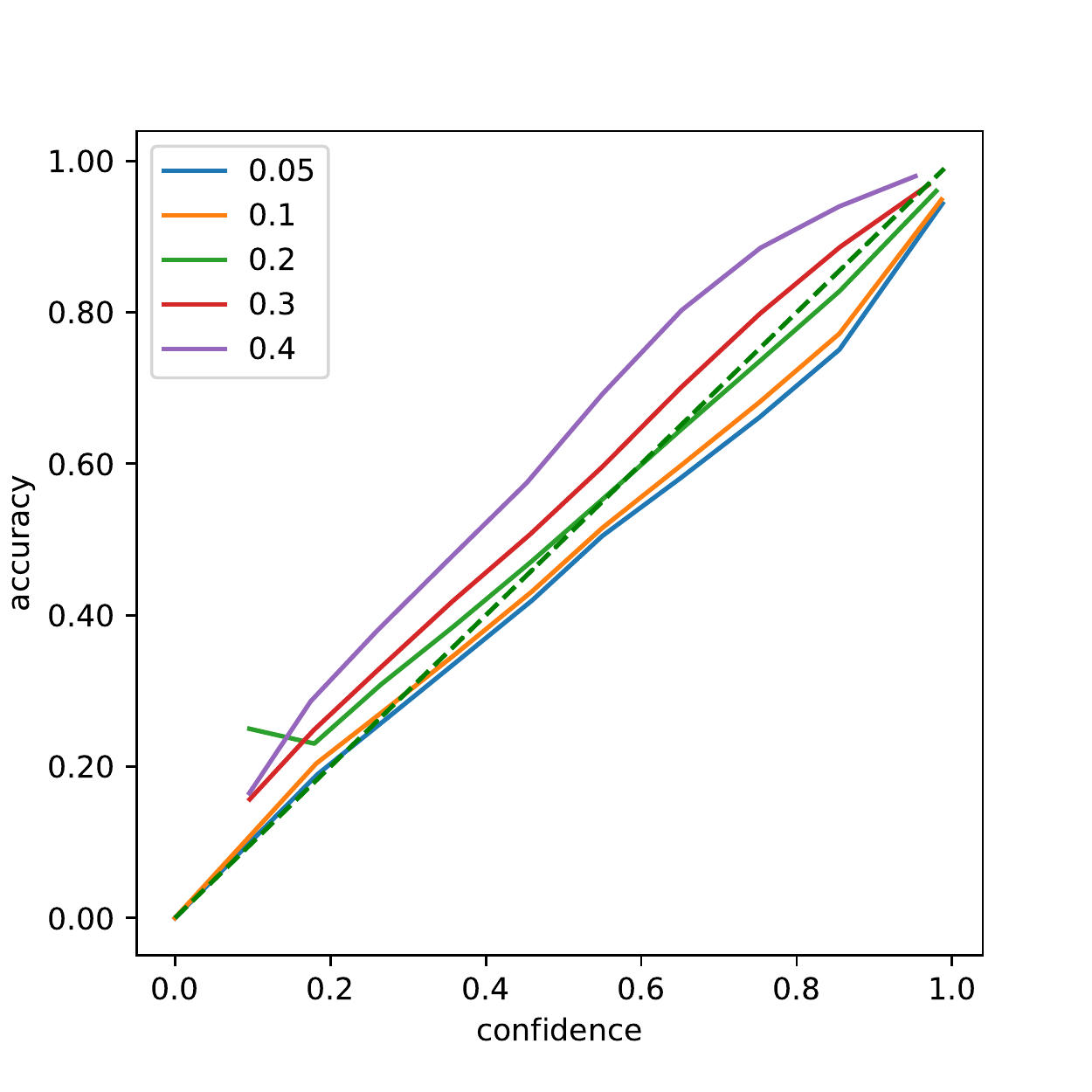}
        \caption{MC-Dropout - Reliability Plot}
        \label{figure:dp_calib}
    \end{minipage}	
\end{figure}	

\subsection{Monte Carlo Dropout}

Models have been evaluated for drop probabilities $p$ of 0.05, 0.1, 0.2, 0.3 and 0.4. Dropout layers were included after each encoder and decoder block. Each model is evaluated with varying number of forward passes ranging from 5 to 30 in increments of 5. 

As seen in Figure \ref{figure:dp_per}, a general trend is that task metrics improve with increasing number of forward passes with the extent being more significant for higher dropout values. Increasing the number of forward passes in a way can be seen as averaging over multiple predictions hence is similar to ensembling which explains this behavior. 

Performance metrics are best for $p = 0.1$ and there is a significant drop in performance for $p = 0.4$ as seen in Figure \ref{figure:dp_per}. In Figure \ref{figure:dp_unc} it can be seen that the entropy value increase with $p$. However, for negative log-likelihood, the lowest value is for $p = 0.2$ but when we see the class level uncertainties in Figure \ref{figure:dp_nll_class} we can see that for most classes, NLL decreases with the increase in $p$. However, this is not the case for classes road and car, as these contain the highest number of points and hence the overall trend differs from the class-wise trend.

Similar to ensembles we can see that in Figures \ref{figure:dp_ep_class} and \ref{figure:dp_nll_class} NLL decreases with the \# of forward passes, while entropy slightly increases. The area of the accuracy vs confidence curve increases with the dropout value is highest for 0.4.

Looking at the calibration plot in Figures \ref{figure:dp_calib} and \ref{figure:dp_ece}, it can be seen that the models seems to be over-confident for lower value of $p$ and gets under-confident for higher values of $p$. The network is best calibrated for $p = 0.2$.

\begin{figure}[t!]
    \centering
    \includegraphics[scale=0.3]{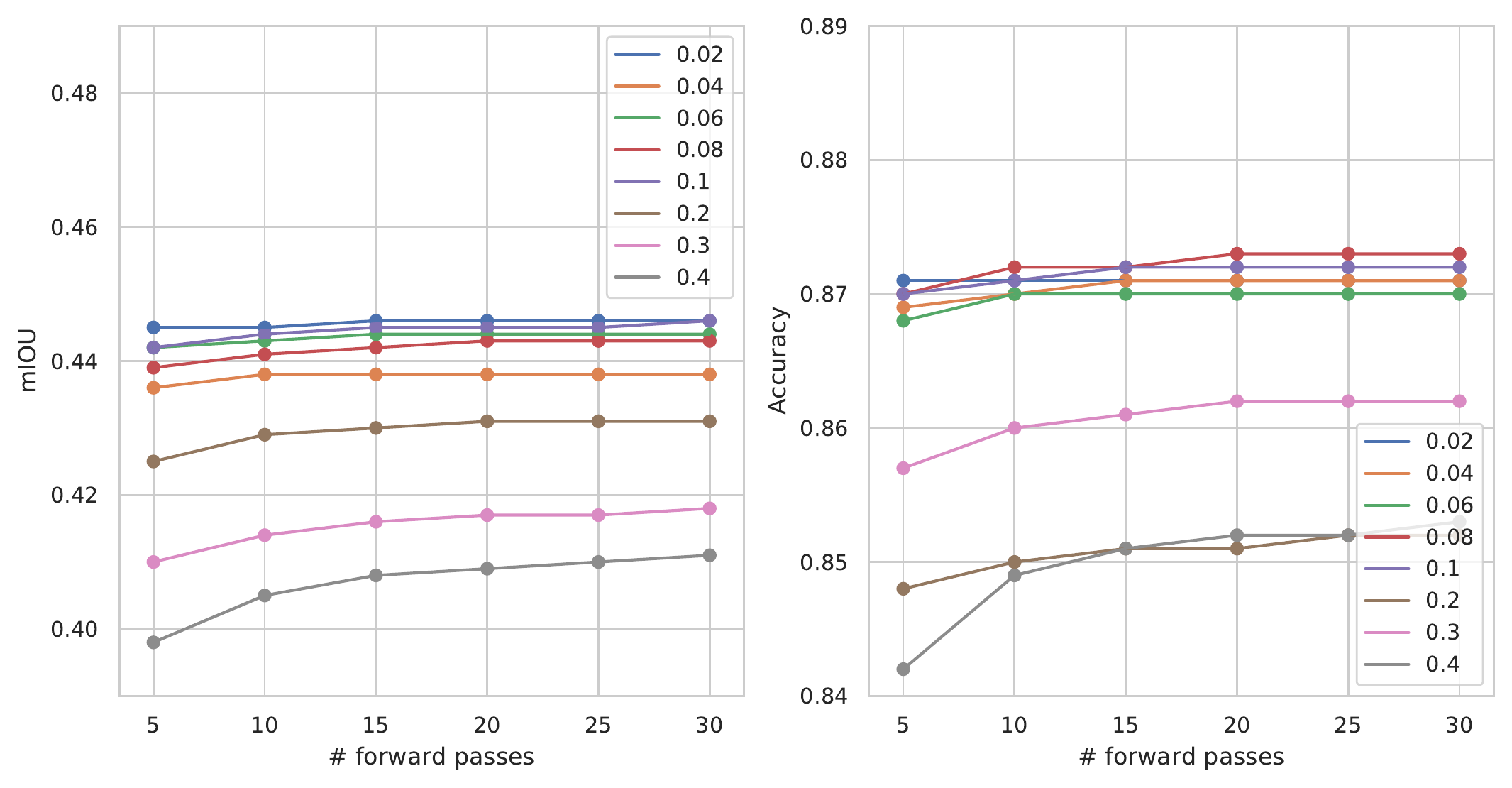}
    \caption{MC-DropConnect - Task Performance vs \# of Forward Passes}
    \label{figure:dc_per}
\end{figure}

\begin{figure}[t!]
    \centering
    \includegraphics[scale=0.30]{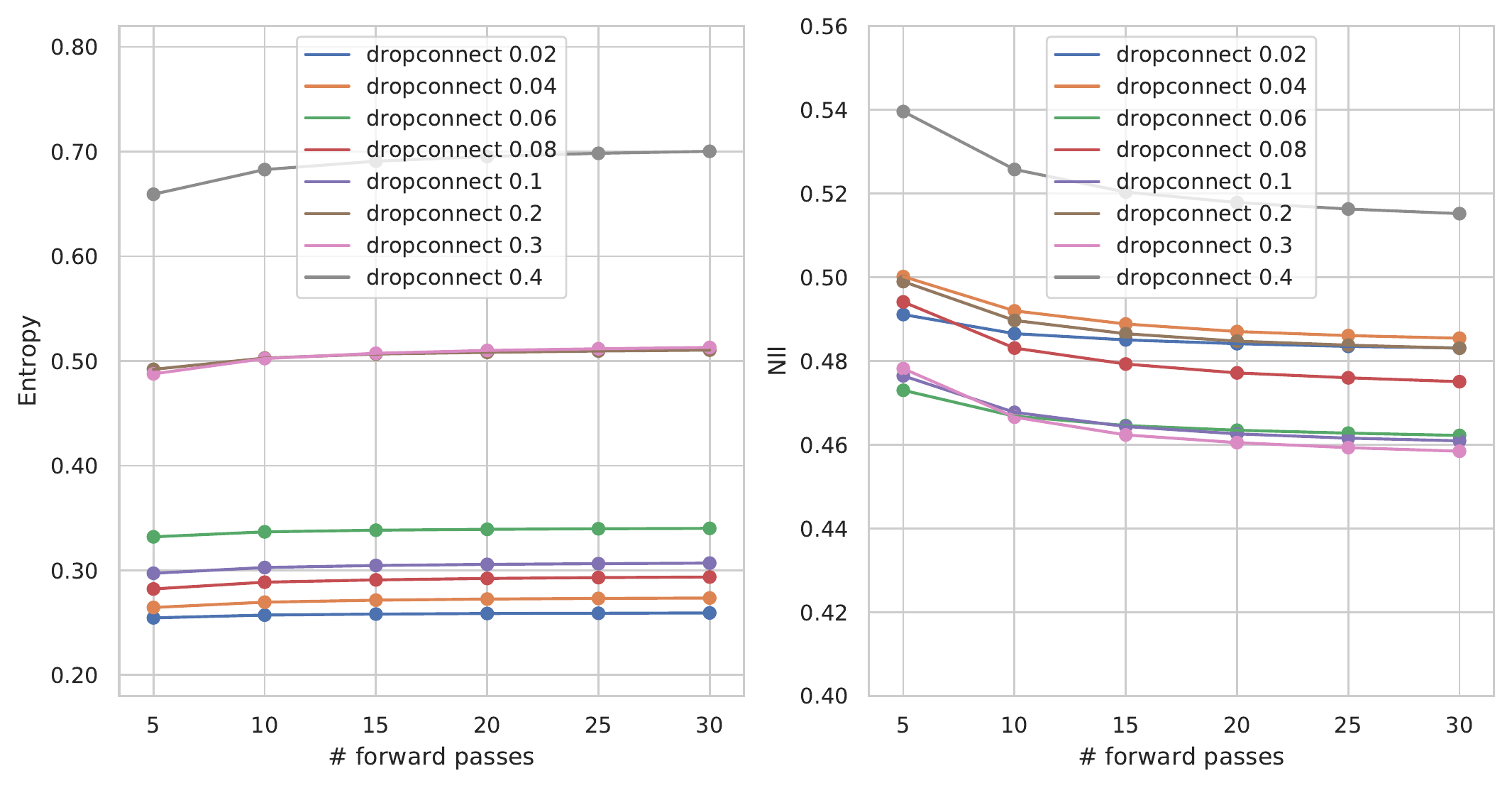}
    \caption{MC-DropConnect - Uncertainty vs \# of Forward Passes}
    \label{figure:dc_unc}
\end{figure}

\subsection{Monte Carlo DropConnect}

Models have been evaluated for varying drop probabilities $p$ values of 0.02, 0.04, 0.06, 0.08, 0.1, 0.2, 0.3 and 0.4.  DropConnect Convolutional layers were implemented and convolution layers were replaced with these layers. Similar to dropout evaluation, each model is evaluated with varying number of forward passes ranging from 5 to 30 in increments of 5. 

In general, MC DropConnect results follows a similar trend as MC Dropout. Metrics improve with the increasing number of forward passes with the extent being more significant for higher values of $p$.

The performance metrics is best for the lowest value of $p = 0.02$ as seen in Figure \ref{figure:dc_per}, however, by increasing the number of forward passes the same performance is achieved for $p = 0.1$. In general performance is severely affected when using DropConnect. In Figure \ref{figure:dc_unc} it can be seen that the entropy values increase with the $p$. The lowest NLL was achieved at $p = 0.3$ with 30 forward passes. However, in Figure \ref{figure:dc_nll_class}, we can see that the per-class NLL uncertainty decreases with increasing $p$ value except for car and road, hence the overall trend differs from class-wise trend.

Figure \ref{figure:dc_ep_class} and \ref{figure:dc_nll_class} shows patterns similar to ensembles and dropout. The area of the accuracy vs confidence curve increases with the $P$ and number of forward passes and is highest for $p = 0.3$.

\begin{figure}[t!]
    \begin{minipage}[t]{0.2\textwidth}	
        \centering
        \includegraphics[scale=0.3]{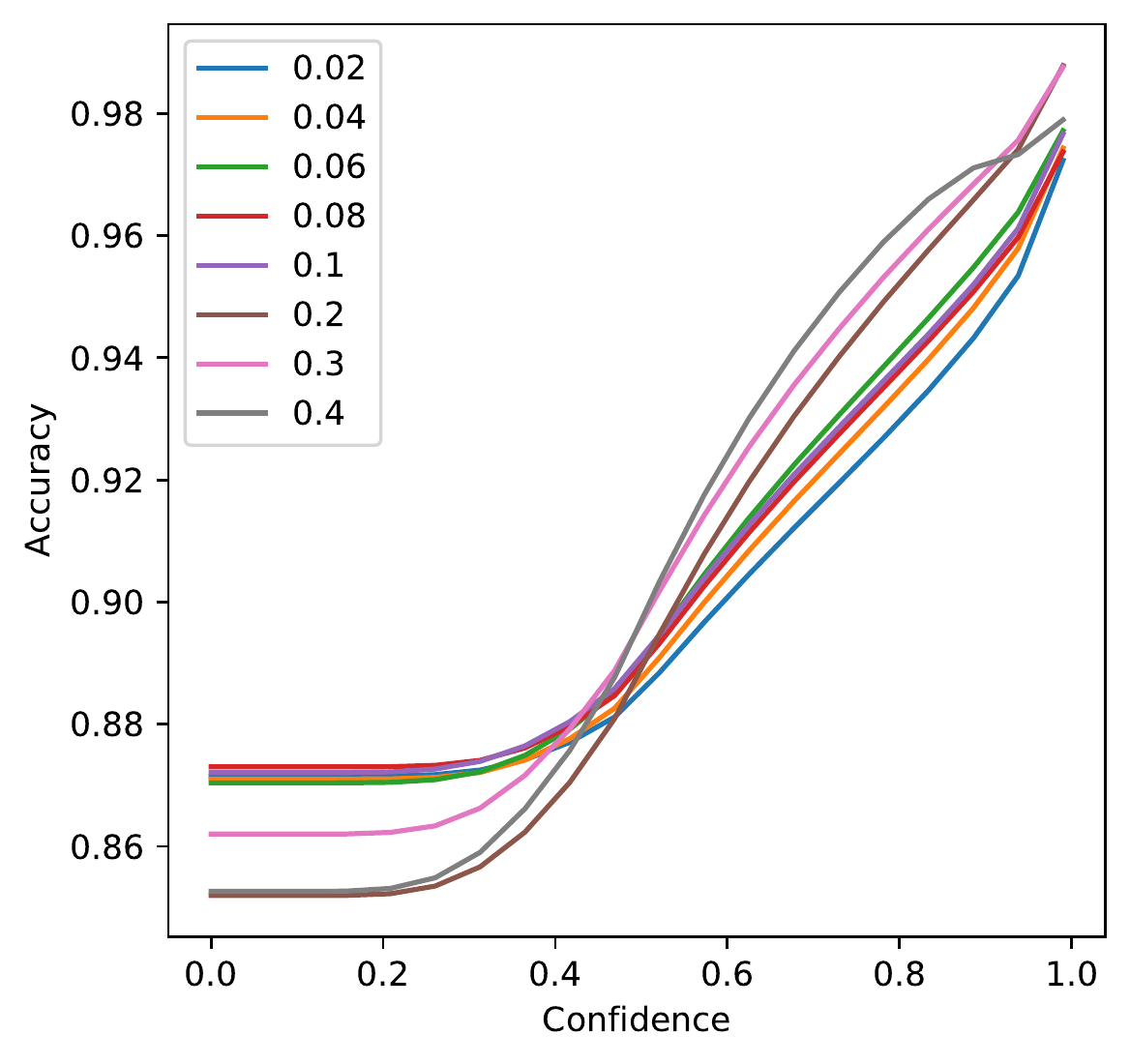}
        \caption{MC-DropConnect - Accuracy vs Confidence}
        \label{figure:dc_acp}
    \end{minipage} \hfill
    \begin{minipage}[t]{0.2\textwidth}	
        \centering
        \includegraphics[scale=0.3]{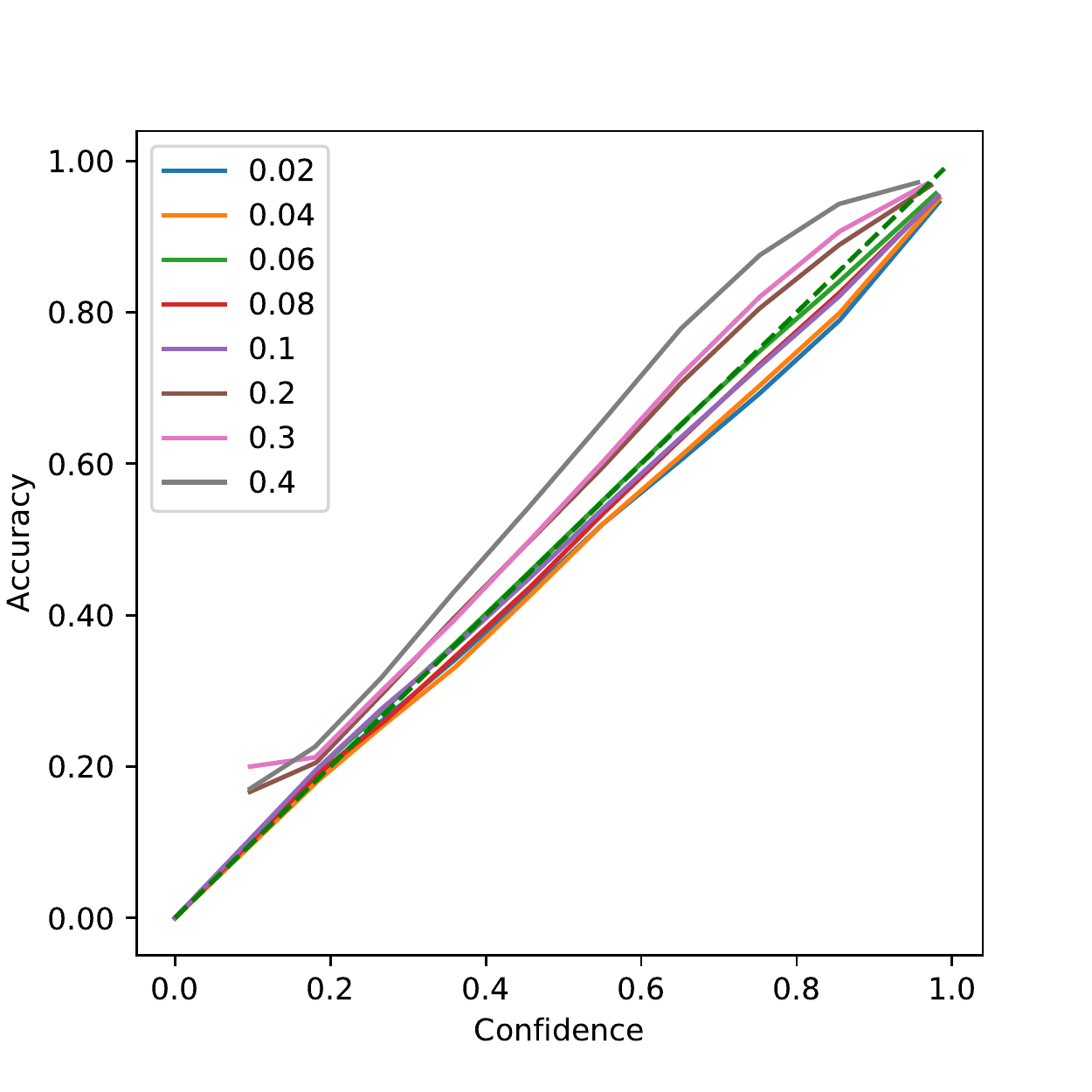}
        \caption{MC-DropConnect - Reliability Plot}
        \label{figure:dc_calib}
    \end{minipage}	
\end{figure}

Looking at the calibration plots in Figure \ref{figure:dc_calib} it can be seen that the model is over-confident for lower values of $p$ and starts to get under-confident  for increasing $p$.
 
%

\section{Conclusions and Future Work}

Our results show that Deep-ensembles outperforms MC Dropout and MC DropConnect in every aspect of our evaluation, followed by MC Dropout and then by MC DropConnect. This is consistent with the literature. For MC-Dropout and MC-Dropconnect, the drop probability $p$ and the number of forward passes needs to be carefully tuned to obtain the best performance, while a Deep Ensemble is simpler to use, as it provides the best uncertainty, and only needs 8-10 models in the ensemble to saturate most metrics.

We also find that higher values of $p$ significantly benefit from increasing the number of forward passes. However this increases the time required for a single prediction. A better way to sample dropout masks can be designed during test time so that the network takes into account the previous dropout masks and the new sampled dropout masks differ significantly from one another rather than just random sampling based on a Bernoulli distribution.

Overall we believe that our results can guide the development of Bayesian Neural Networks for point cloud segmentation, which we expect can improve the safety and decision making of many applications that rely on this kind of perception, including autonomous driving and crop harvesting.

As future work we wish to evaluate out of distribution detection, and consider models other than DarkNet21Seg for this task.

\clearpage
\bibliography{uncertainty_estimation}
\bibliographystyle{icml2020}

\clearpage
\onecolumn
\appendix
\section{Calibration vs Drop Probabilities}

\begin{figure}[h!]
    \centering
    \begin{subfigure}{0.48\textwidth}
        \centering
        \includegraphics[scale=0.5]{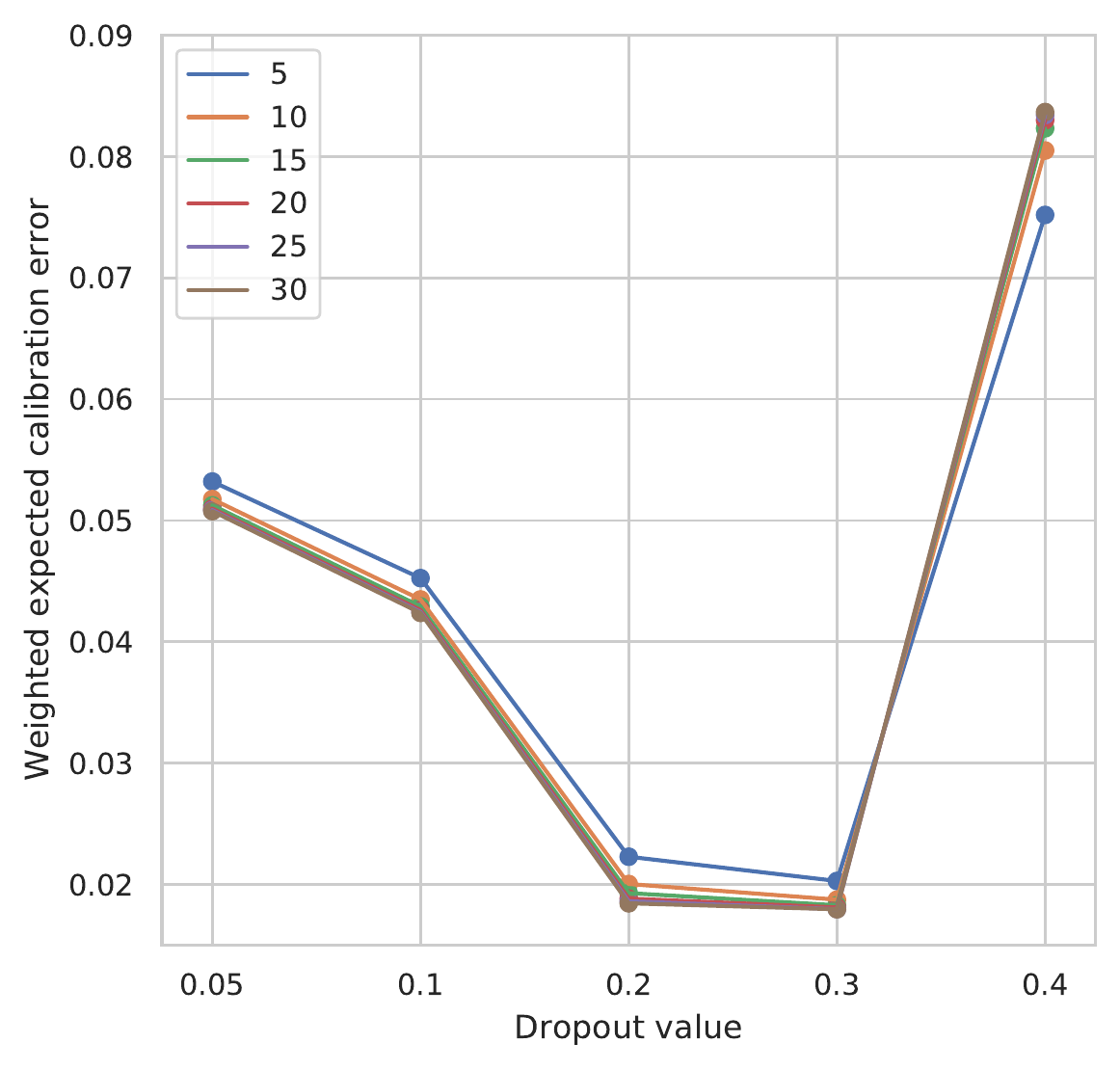}
        \caption{MC-Dropout}
        \label{figure:dp_ece}
    \end{subfigure}
    \begin{subfigure}{0.48\textwidth}
        \centering
        \includegraphics[scale=0.5]{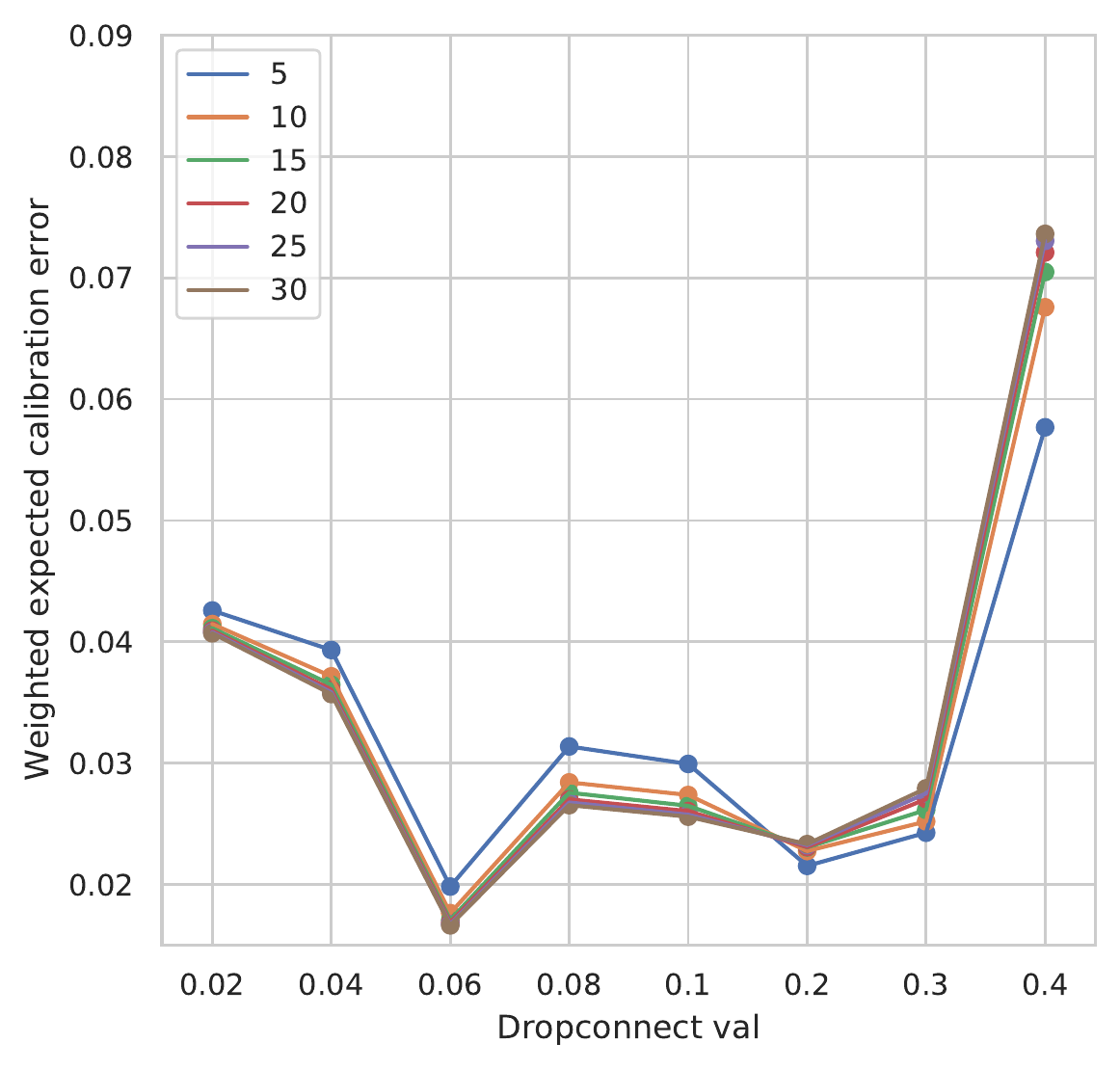}
        \caption{MC-DropConnect}
        \label{figure:dc_ece}
    \end{subfigure}
    \caption{Calibration Metrics vs Drop Probability for MC-Dropout and MC-DropConnect Models}
\end{figure}

This section presents two plots that did not fit into the main paper, but are part of our main results regarding calibration as the drop probability $p$ is varied. These are shown in Figures \ref{figure:dp_ece} and Figure \ref{figure:dc_ece}.

\section{Per-class Entropy}

\begin{figure}[h]
    \centering
    \begin{subfigure}{0.48\textwidth}
        \centering
        \includegraphics[width=\textwidth]{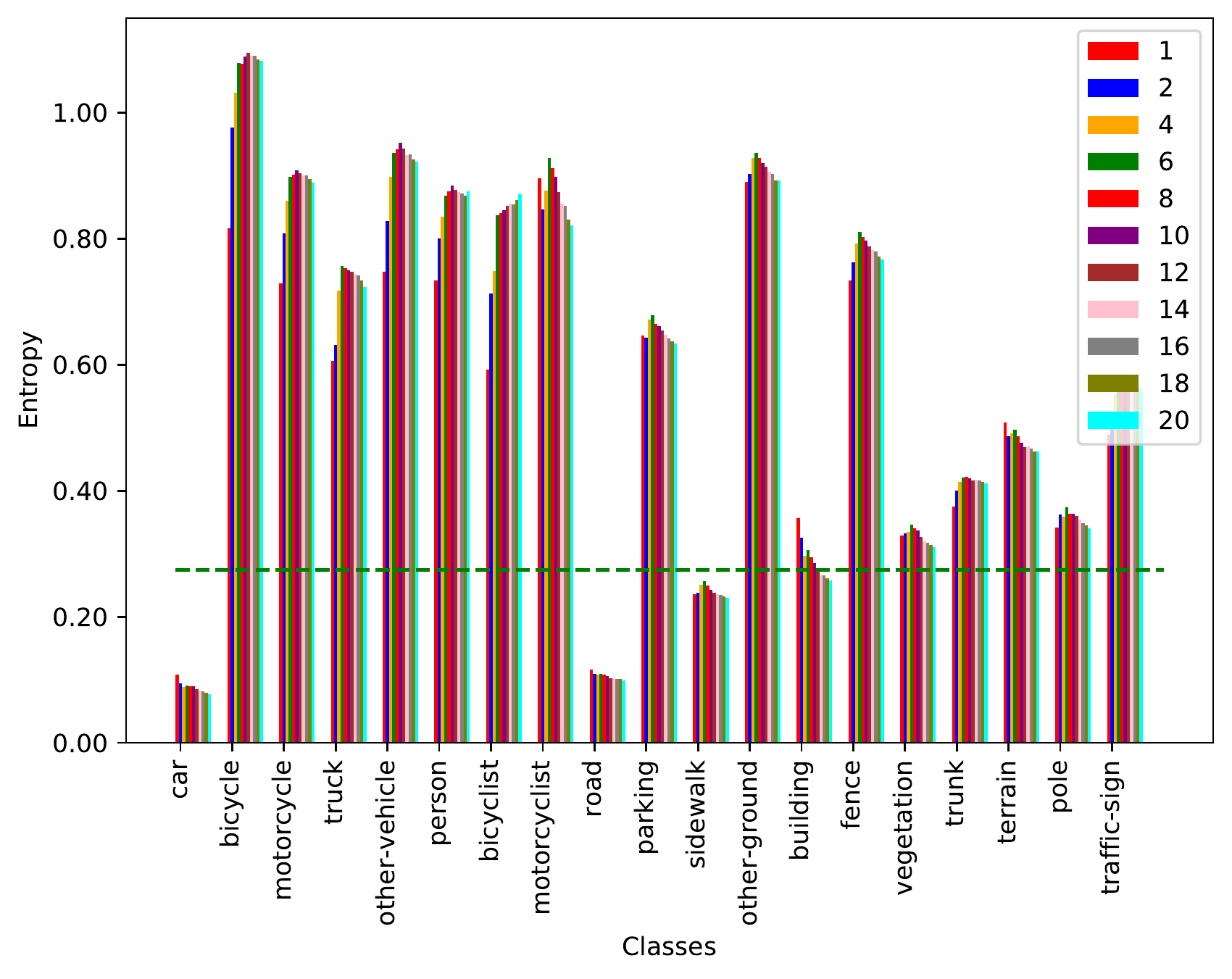}
        \caption{Deep Ensembles}
        \label{figure:ens_ep_class}
    \end{subfigure}
    \begin{subfigure}{0.48\textwidth}
        \centering
        \includegraphics[width=\textwidth]{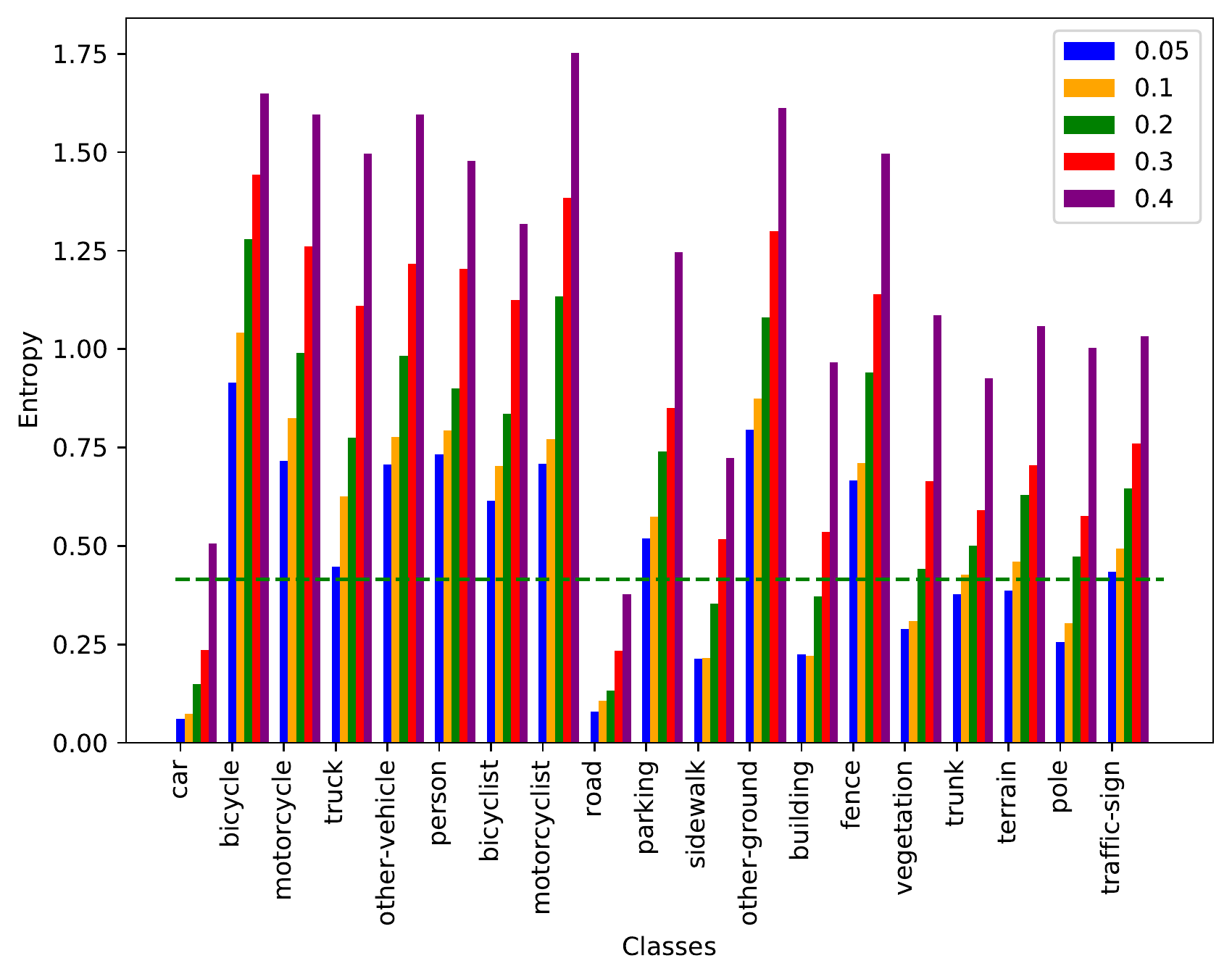}
        \caption{MC-Dropout}
        \label{figure:dp_ep_class}
    \end{subfigure}
    \begin{subfigure}{0.48\textwidth}
        \centering
        \includegraphics[width=\textwidth]{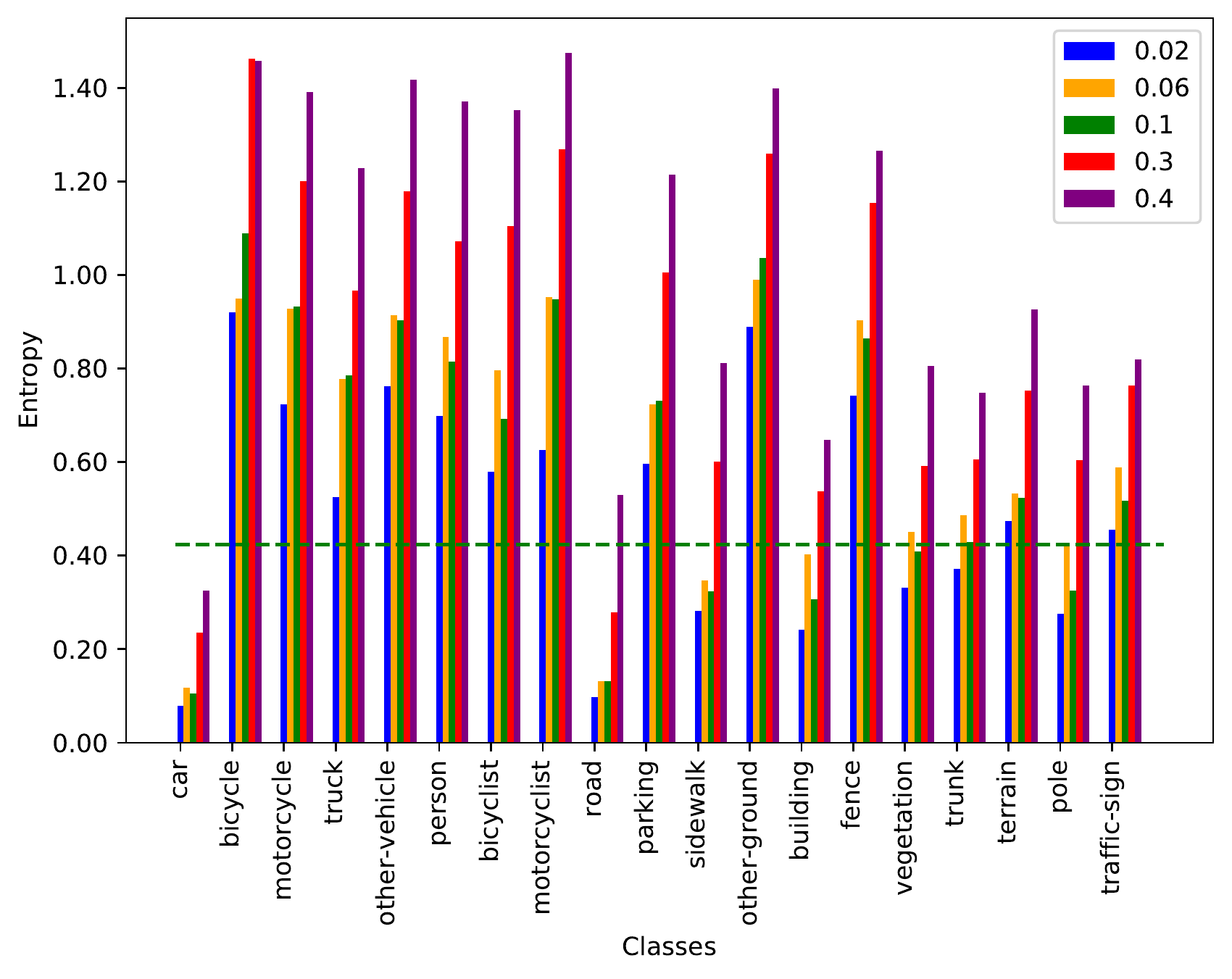}
        \caption{MC-DropConnect}
        \label{figure:dc_ep_class}
    \end{subfigure}

    \caption{Comparison of class-wise Entropy}
    \label{figure:ep_class_all}
\end{figure}

In this section we include additional results of predictive entropy for each class and for each method. Entropy is directly related to the uncertainty predicted by the model, which should indicate if some classes are overall more uncertain than others. We present these results in Figure \ref{figure:ep_class_all}.

Deep Ensembles provides more consistent entropy values, with it increasing or decreasing depending on the class as the number of ensembles is varied. Entropy values are compatible with the frequency of each class, classes with more data points having lower uncertainty than classes with less data points. This is particularly noticeable for the car and road classes, which are the majority in the SemanticKITTI dataset, while bicycle, motorcycle and bicyclist have the highest entropy due to the confusion between these classes and the low number of samples.

MC Dropout and MC Dropconnect are overall more uncertain than Deep Ensembles, with more variation between entropy values as the drop probability $p$ is varied, which makes this parameter harder to tune. There are similar relations between entropy values produced by these methods, and the number of samples for each class.

\FloatBarrier
\section{Per-class Negative Log-Likelihood}

\begin{figure}[h]
    \centering
    \begin{subfigure}{0.48\textwidth}
        \centering
        \includegraphics[width=\textwidth]{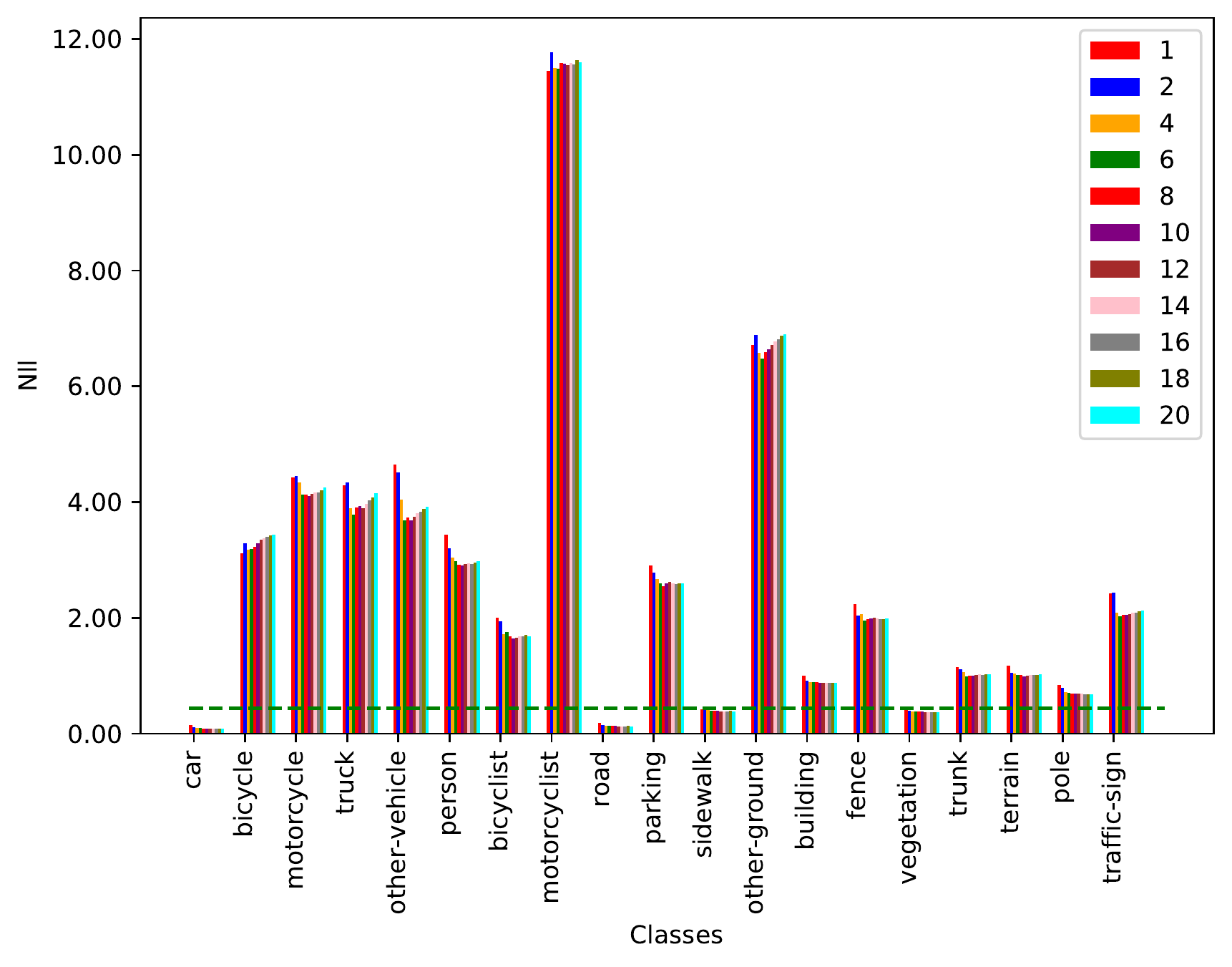}
        \caption{Deep Ensembles}
        \label{figure:ens_nll_class}
    \end{subfigure}
    \begin{subfigure}{0.48\textwidth}
        \centering
        \includegraphics[width=\textwidth]{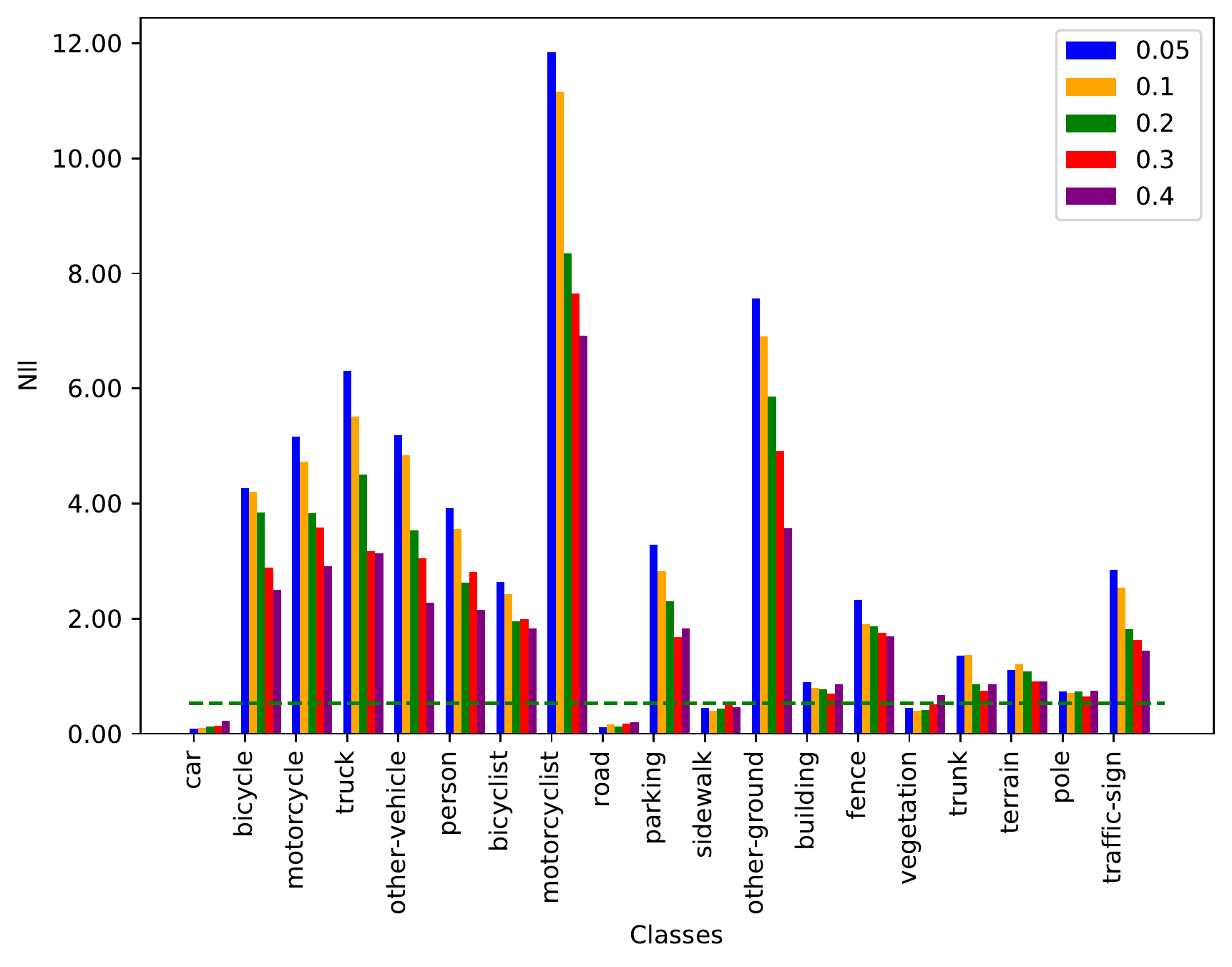}
        \caption{MC-Dropout}
        \label{figure:dp_nll_class}
    \end{subfigure}
    \begin{subfigure}{0.48\textwidth}
        \centering
        \includegraphics[width=\textwidth]{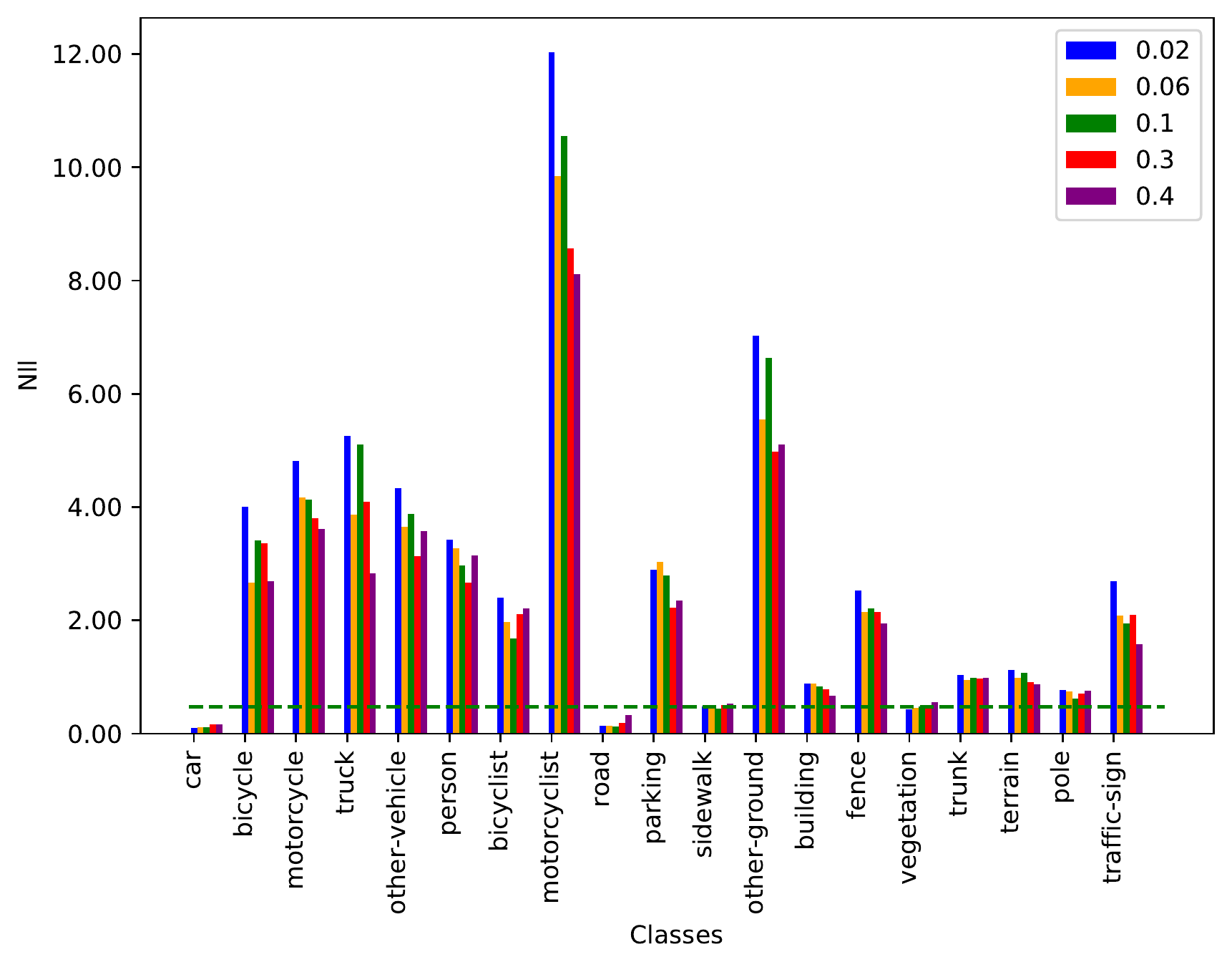}
        \caption{MC-DropConnect}
        \label{figure:dc_nll_class}
    \end{subfigure}
    
    \caption{Comparison of class-wise Negative Log-Likelihood}
    \label{figure:nll_class}
\end{figure}

In this section we include additional results of the negative log-likelihood for each class and for each method. These results are meant to disentangle effects of the aggregated loss versus its per class components. Results are shown in Figure \ref{figure:nll_class}.

All three uncertainty methods show very similar patterns regarding negative log-likelihood, with Deep Ensembles having minor variations (increase or decreases) as the number of samples is varied, and MC Dropout and MC DropConnect having larger variations across different values of the drop probability $p$.

As expected, lower NLL values are produced for classes with more samples per class, like car and road, and higher NLL values are seen for highly uncertain or underrepresented classes, such as motorcyclist, other-ground, and different kinds of vehicles. We believe that this indicated that the uncertainty methods are not introducing additional biases in the model, as the three methods produce very similar results when NLL is separated per class.

\FloatBarrier
\section{Per-class comparison of Task Performance}

In this section we present additional results for per-class task performance, namely mean IoU and mean per-pixel accuracy, as the number of ensembles or drop probability $p$ is varied. These results are presented in Table \ref{table:darknet_perf_iou} for mIoU, and Table \ref{table:darknet_perf_acc} for per-pixel accuracy.

Class-wise IoU shows that Deep Ensembles outperforms all other methods, and the original model without uncertainty quantification, for most of the classes, in particular MC DropConnect seems to perform best for the Fence class.

Per-pixel accuracy results are more mixed, with Deep Ensembles still performing better overall, and MC Dropout outperforming Deep Ensembles for some classes, particularly Car, Truck, Other-vehicle, and Other-ground. MC DropConnect again has the best performance for the Fence class.

\begin{table*}[htb]
    \centering
    \resizebox{\textwidth}{!}{%
        \begin{tabular}{llllllllllllllllllllll}
            \toprule
            \begin{turn}{90} Uncertainty method\end{turn} &
            \begin{turn}{90} Value\end{turn} &
            \begin{turn}{90} Mean IoU\end{turn}        & \begin{turn}{90}Car\end{turn}              & \begin{turn}{90}Bicycle\end{turn}               & \begin{turn}{90}Motorcycle\end{turn}               & \begin{turn}{90}Truck\end{turn}              & \begin{turn}{90}Other-vehicle\end{turn}               & \begin{turn}{90}Person\end{turn}               & \begin{turn}{90}Bicyclist\end{turn}                                & \begin{turn}{90}Motorcyclist\end{turn}   & \begin{turn}{90}Road\end{turn}               & \begin{turn}{90}Parking\end{turn}              & \begin{turn}{90}Sidewalk\end{turn}              & \begin{turn}{90}Other-ground\end{turn}             & \begin{turn}{90}Building\end{turn}              & \begin{turn}{90}Fence\end{turn}              & \begin{turn}{90}Vegetation\end{turn}              & \begin{turn}{90}Trunk\end{turn}              & \begin{turn}{90}Terrain\end{turn}              & \begin{turn}{90}Pole\end{turn}              & \begin{turn}{90}Traffic-sign\end{turn}              \\
            \midrule
            & & 
            &                 &                 &                 &                 &                 &                 &                 &     &                 &                 &                 &                &                 &                 &                 &                 &                 &                 & \\
            None & NA & 0.449 & 0.845 & 0.156 & 0.301 & 0.170 & 0.275 & 0.337 & 0.508 & 0.000 & 0.923 & 0.440 & 0.781 & 0.000 & 0.766 & 0.476 & 0.801 & 0.414 & 0.719 & 0.315 & 0.302\\
            \midrule
            Deep ensembles & 2 & 0.469 & 0.825 & 0.214 & 0.383 & 0.180 & 0.227 & 0.379 & 0.514 & 0.000 & 0.935 & 0.464 & 0.805 & 0.000 & 0.791 & 0.465 & 0.810 & 0.468 & 0.718 & 0.364 & 0.362  \\
            Deep ensembles & 4 & 0.473 & 0.821 & 0.238 & 0.393 & 0.145 & 0.196 & 0.402 & 0.541 & 0.000 & 0.938 & 0.500 & 0.810 & 0.000 & 0.796 & 0.466 & 0.816 & 0.484 & 0.722 & 0.360 & 0.366 \\
            Deep ensembles & 6 & 0.480 & 0.826 & 0.245 & 0.407 & 0.172 & 0.210 & 0.416 & 0.546 & 0.000 & 0.938 & 0.497 & 0.811 & 0.000 & 0.800 & 0.480 & 0.818 & 0.485 & 0.727 & 0.374 & 0.370 \\
            Deep ensembles & 8 & 0.484 & 0.826 & 0.258 & 0.421 & 0.158 & 0.211 & 0.419 & 0.554 & 0.000 & 0.940 & \textbf{0.516} & 0.813 & 0.001 & 0.802 & 0.483 & 0.818 & 0.495 & 0.721 & 0.384 & 0.382 \\
            Deep ensembles & 10 & \textbf{0.485} & 0.826 & 0.264 & 0.427 & 0.148 & 0.207 & 0.424 & \textbf{0.556} & 0.000 & 0.939 & 0.508 & 0.812 & 0.001 & 0.804 & 0.485 & 0.819 & 0.498 & 0.727 & 0.391 & 0.385 \\
            Deep ensembles & 12 & \textbf{0.485} & 0.820 & 0.267 & \textbf{0.428} & 0.143 & 0.191 & 0.427 & \textbf{0.556} & 0.000 & \textbf{0.941} & 0.507 & 0.814 & 0.001 & 0.804 & 0.473 & 0.822 & 0.503 & 0.730 & \textbf{0.392} & 0.388 \\
            Deep ensembles & 14 & 0.484 & 0.816 & 0.267 & 0.426 & 0.136 & 0.185 & 0.427 & 0.555 & 0.000 & \textbf{0.941} & 0.513 & 0.815 & 0.000 & 0.805 & 0.467 & 0.822 & 0.506 & 0.730 & \textbf{0.392} & 0.389 \\
            Deep ensembles & 16 & 0.484 & 0.818 & 0.266 & 0.425 & 0.138 & 0.186 & \textbf{0.429} & 0.553 & 0.000 & \textbf{0.941} & 0.512 & \textbf{0.816} & 0.000 & 0.805 & 0.472 & \textbf{0.823} & 0.505 & \textbf{0.731} & 0.389 & \textbf{0.390} \\
            Deep ensembles & 18 & 0.484 & 0.819 & 0.268 & 0.427 & 0.131 & 0.182 & 0.427 & 0.551 & 0.000 & 0.940 & 0.510 & \textbf{0.816} & 0.001 & \textbf{0.806} & 0.474 & \textbf{0.823} & 0.506 & 0.729 & 0.390 & 0.389 \\
            Deep ensembles & 20 & 0.484 & 0.820 & \textbf{0.269} & 0.427 & 0.128 & 0.186 & 0.427 & 0.549 & 0.000 & \textbf{0.941} & 0.508 & \textbf{0.816} & 0.000 & \textbf{0.806} & 0.476 & \textbf{0.823} & \textbf{0.507} & 0.730 & 0.389 & 0.388 \\
            \midrule
            MC-Dropout & 0.05 & 0.461 & 0.835 & 0.219 & 0.373 & \textbf{0.192} & 0.247 & 0.362 & 0.447 & 0.000 & 0.932 & 0.460 & 0.794 & 0.002 & 0.791 & 0.448 & 0.804 & 0.462 & 0.718 & 0.337 & 0.330 \\
            MC-Dropout & 0.1 & 0.470 & 0.845 & 0.233 & 0.383 & 0.191 & 0.258 & 0.378 & 0.491 & 0.000 & 0.934 & 0.476 & 0.806 & 0.001 & 0.790 & 0.452 & 0.810 & 0.465 & 0.722 & 0.348 & 0.357 \\
            MC-Dropout & 0.2 & 0.455 & \textbf{0.850} & 0.211 & 0.372 & 0.179 & 0.283 & 0.349 & 0.407 & 0.000 & 0.934 & 0.452 & 0.798 & 0.001 & 0.775 & 0.447 & 0.803 & 0.402 & 0.719 & 0.338 & 0.334 \\
            MC-Dropout & 0.3 & 0.452 & 0.835 & 0.188 & 0.401 & 0.175 & \textbf{0.319} & 0.356 & 0.440 & 0.000 & 0.929 & 0.396 & 0.775 & 0.009 & 0.776 & 0.467 & 0.796 & 0.389 & 0.732 & 0.293 & 0.322 \\
            MC-Dropout & 0.4 & 0.431 & 0.833 & 0.171 & 0.307 & 0.122 & 0.292 & 0.289 & 0.397 & 0.000 & 0.924 & 0.428 & 0.778 & \textbf{0.022} & 0.742 & 0.439 & 0.778 & 0.390 & 0.722 & 0.286 & 0.279 \\
            \midrule
            MC-DropConnect & 0.02 & 0.444 & 0.834 & 0.213 & 0.362 & 0.085 & 0.159 & 0.357 & 0.449 & 0.000 & 0.934 & 0.430 & 0.793 & 0.001 & 0.787 & \textbf{0.512} & 0.793 & 0.416 & 0.686 & 0.322 & 0.309 \\
            MC-DropConnect & 0.04 &  0.436 & 0.832 & 0.204 & 0.341 & 0.126 & 0.085 & 0.341 & 0.493 & 0.000 & 0.927 & 0.404 & 0.776 & 0.001 & 0.766 & 0.383 & 0.804 & 0.443 & 0.711 & 0.332 & 0.316 \\
            MC-DropConnect & 0.06 & 0.442 & 0.815 & 0.160 & 0.350 & 0.110 & 0.195 & 0.335 & 0.472 & 0.000 & 0.930 & 0.397 & 0.773 & 0.006 & 0.780 & 0.510 & 0.799 & 0.410 & 0.719 & 0.316 & 0.315 \\
            MC-DropConnect & 0.08 & 0.441 & 0.827 & 0.226 & 0.350 & 0.019 & 0.168 & 0.344 & 0.483 & 0.000 & 0.926 & 0.428 & 0.784 & 0.000 & 0.772 & 0.423 & 0.806 & 0.438 & 0.718 & 0.341 & 0.330 \\
            MC-DropConnect & 0.1 & 0.442 & 0.839 & 0.198 & 0.321 & 0.110 & 0.266 & 0.324 & 0.454 & 0.000 & 0.931 & 0.428 & 0.785 & 0.001 & 0.774 & 0.437 & 0.797 & 0.418 & 0.725 & 0.302 & 0.297 \\
            MC-DropConnect & 0.2 & 0.425 & 0.838 & 0.189 & 0.339 & 0.100 & 0.269 & 0.308 & 0.451 & 0.000 & 0.920 & 0.278 & 0.711 & 0.017 & 0.750 & 0.438 & 0.782 & 0.410 & 0.681 & 0.293 & 0.305 \\
            MC-DropConnect & 0.3 & 0.415 & 0.798 & 0.153 & 0.253 & 0.129 & 0.243 & 0.277 & 0.358 & 0.000 & 0.921 & 0.378 & 0.770 & 0.001 & 0.752 & 0.391 & 0.788 & 0.394 & 0.707 & 0.272 & 0.293 \\
            MC-DropConnect & 0.4 & 0.421 & 0.823 & 0.175 & 0.309 & 0.116 & 0.126 & 0.283 & 0.417 & 0.000 & 0.921 & 0.388 & 0.771 & 0.001 & 0.755 & 0.494 & 0.792 & 0.405 & 0.719 & 0.270 & 0.241\\
            \bottomrule
    \end{tabular}}
    \caption{Class-wise IOU on 08 sequence of SemanticKITTI. The None value indicates the baseline model without any uncertainty quantification.}
    \label{table:darknet_perf_iou}
\end{table*}

\begin{table*}[htb]
    \centering
    \resizebox{\textwidth}{!}{%
        \begin{tabular}{llllllllllllllllllllll}
            \toprule
            \begin{turn}{90} Uncertainty method\end{turn} &
            \begin{turn}{90} Value\end{turn} &
            \begin{turn}{90} Mean Accuracy\end{turn}        & \begin{turn}{90}Car\end{turn}              & \begin{turn}{90}Bicycle\end{turn}               & \begin{turn}{90}Motorcycle\end{turn}               & \begin{turn}{90}Truck\end{turn}              & \begin{turn}{90}Other-vehicle\end{turn}               & \begin{turn}{90}Person\end{turn}               & \begin{turn}{90}Bicyclist\end{turn}                                & \begin{turn}{90}Motorcyclist\end{turn}   & \begin{turn}{90}Road\end{turn}               & \begin{turn}{90}Parking\end{turn}              & \begin{turn}{90}Sidewalk\end{turn}              & \begin{turn}{90}Other-ground\end{turn}             & \begin{turn}{90}Building\end{turn}              & \begin{turn}{90}Fence\end{turn}              & \begin{turn}{90}Vegetation\end{turn}              & \begin{turn}{90}Trunk\end{turn}              & \begin{turn}{90}Terrain\end{turn}              & \begin{turn}{90}Pole\end{turn}              & \begin{turn}{90}Traffic-sign\end{turn}              \\
            \midrule 
            & & 
            &                 &                 &                 &                 &                 &                 &                 &     &                 &                 &                 &                &                 &                 &                 &                 &                 &                 & \\
            None & NA & 0.869 & \textbf{0.870} & 0.171 & 0.384 & 0.221 & 0.532 & 0.433 & 0.595 & 0.000 & \textbf{0.983} & 0.682 & 0.860 & 0.000 & \textbf{0.936} & 0.553 & 0.888 & 0.474 & 0.872 & 0.350 & 0.451 \\
            \midrule
            Deep ensembles & 2 & 0.879 & 0.837 & 0.250 & 0.510 & 0.378 & 0.630 & 0.494 & 0.620 & 0.000 & 0.979 & 0.723 & 0.884 & 0.001 & 0.925 & 0.586 & 0.893 & 0.543 & 0.861 & 0.409 & 0.587 \\
            
            Deep ensembles & 4 & 0.882 & 0.831 & 0.283 & 0.530 & 0.504 & 0.633 & 0.528 & 0.632 & 0.000 & 0.980 & 0.759 & 0.888 & 0.001 & 0.919 & 0.614 & 0.895 & 0.559 & 0.864 & 0.399 & 0.553 \\
            
            Deep ensembles & 6 & 0.884 & 0.836 & 0.297 & 0.533 & 0.569 & 0.657 & 0.551 & 0.649 & 0.000 & 0.981 & 0.753 & 0.885 & 0.002 & 0.925 & 0.615 & 0.897 & 0.556 & 0.866 & 0.417 & 0.578 \\
            
            Deep ensembles & 8 & 0.885 & 0.835 & 0.317 & 0.560 & 0.604 & 0.666 & 0.561 & 0.653 & 0.000 & 0.982 & 0.749 & 0.887 & 0.002 & 0.925 & 0.624 & 0.894 & 0.571 & 0.867 & 0.427 & 0.606 \\
            
            Deep ensembles & 10 & 0.886 & 0.835 & 0.329 & 0.568 & 0.626 & 0.674 & 0.567 & 0.657 & 0.000 & 0.982 & 0.754 & 0.885 & 0.003 & 0.925 & 0.624 & 0.898 & 0.575 & 0.864 & 0.435 & 0.622 \\
            
            Deep ensembles & 12 & 0.886 & 0.829 & 0.339 & 0.576 & 0.647 & 0.675 & 0.573 & 0.659 & 0.000 & 0.982 & \textbf{0.763} & 0.887 & 0.003 & 0.922 & 0.624 & 0.899 & 0.582 & 0.868 & \textbf{0.437} & 0.629 \\
            
            Deep ensembles & 14 & 0.886 & 0.824 & 0.342 & 0.574 & 0.655 & 0.680 & 0.572 & 0.661 & 0.000 & 0.982 & 0.756 & 0.887 & 0.002 & 0.923 & 0.622 & 0.898 & 0.587 & 0.871 & 0.436 & 0.637 \\
            
            Deep ensembles & 16 & \textbf{0.887} & 0.826 & 0.344 & 0.578 & \textbf{0.656} & 0.682 & 0.574 & 0.658 & 0.000 & 0.982 & 0.751 & 0.887 & 0.002 & 0.924 & 0.625 & 0.898 & 0.585 & 0.872 & 0.432 & 0.638 \\
            
            Deep ensembles & 18 & \textbf{0.887} & 0.827 & 0.348 & 0.580 & 0.655 & 0.684 & 0.574 & \textbf{0.662} & 0.000 & 0.982 & 0.751 & 0.887 & 0.003 & 0.924 & 0.626 & 0.897 & \textbf{0.589} & 0.871 & 0.433 & 0.642 \\
            
            Deep ensembles & 20 & \textbf{0.887} & 0.828 & \textbf{0.351} & \textbf{0.583} & 0.654 & \textbf{0.689} & \textbf{0.581} & 0.659 & 0.000 & 0.982 & 0.754 & 0.888 & 0.002 & 0.924 & 0.627 & 0.897 & \textbf{0.589} & 0.873 & 0.431 & \textbf{0.644} \\
            
            \midrule
            MC-Dropout & 0.05 & 0.876 & 0.846 & 0.263 & 0.515 & 0.566 & 0.590 & 0.510 & 0.585 & 0.000 & 0.967 & 0.687 & 0.886 & 0.004 & 0.910 & 0.574 & 0.902 & 0.547 & 0.849 & 0.368 & 0.519 \\
            
            MC-Dropout & 0.1 & 0.879 & 0.857 & 0.316 & 0.523 & 0.468 & 0.626 & 0.501 & 0.620 & 0.000 & \textbf{0.983} & 0.681 & 0.884 & 0.003 & 0.898 & 0.557 & 0.895 & 0.558 & \textbf{0.880} & 0.383 & 0.543 \\
            
            MC-Dropout & 0.2 & 0.874 & 0.868 & 0.266 & 0.472 & 0.484 & 0.526 & 0.413 & 0.476 & 0.000 & 0.972 & 0.585 & \textbf{0.902} & 0.002 & 0.901 & 0.588 & 0.891 & 0.441 & 0.879 & 0.380 & 0.457 \\
            
            MC-Dropout & 0.3 & 0.868 & 0.853 & 0.211 & 0.532 & 0.360 & 0.537 & 0.450 & 0.523 & 0.000 & \textbf{0.983} & 0.470 & 0.899 & 0.016 & 0.905 & 0.579 & 0.906 & 0.421 & 0.855 & 0.319 & 0.454 \\
            
            MC-Dropout & 0.4 & 0.857 & 0.859 & 0.183 & 0.396 & 0.297 & 0.405 & 0.333 & 0.464 & 0.000 & 0.974 & 0.599 & 0.871 & \textbf{0.025} & 0.915 & 0.533 & \textbf{0.914} & 0.422 & 0.835 & 0.318 & 0.355 \\
            
            \midrule
            MC-DropConnect & 0.02 & 0.871 & 0.846 & 0.271 & 0.491 & 0.387 & 0.520 & 0.465 & 0.548 & 0.000 & 0.975 & 0.604 & 0.888 & 0.003 & 0.906 & \textbf{0.654} & 0.885 & 0.471 & 0.847 & 0.352 & 0.460 \\
            
            MC-DropConnect & 0.04 & 0.870 & 0.843 & 0.300 & 0.476 & 0.412 & 0.560 & 0.500 & 0.624 & 0.000 & 0.959 & 0.694 & 0.870 & 0.002 & 0.877 & 0.569 & 0.894 & 0.504 & 0.861 & 0.373 & 0.545 \\
            
            MC-DropConnect & 0.06 & 0.869 & 0.828 & 0.174 & 0.482 & 0.480 & 0.464 & 0.423 & 0.604 & 0.000 & 0.974 & 0.570 & 0.879 & 0.009 & 0.935 & 0.634 & 0.900 & 0.459 & 0.840 & 0.349 & 0.455 \\
            
            MC-DropConnect & 0.08 & 0.872 & 0.838 & 0.298 & 0.553 & 0.268 & 0.571 & 
            0.517 & 0.629 & 0.000 & \textbf{0.983} & 0.574 & 0.863 & 0.001 & 0.914 & 0.591 & 0.893 & 0.510 & 0.846 & 0.377 & 0.547 \\
            
            MC-DropConnect & 0.1 & 0.871 & 0.856 & 0.245 & 0.457 & 0.480 & 0.465 & 0.412 & 0.565 & 0.000 & 0.969 & 0.606 & 0.878 & 0.002 & 0.895 & 0.596 & 0.910 & 0.469 & 0.842 & 0.327 & 0.396 \\
            
            MC-DropConnect & 0.2 & 0.849 & 0.863 & 0.215 & 0.438 & 0.376 & 0.497 & 0.381 & 0.505 & 0.000 & 0.965 & 0.338 & 0.872 & 0.021 & 0.908 & 0.556 & 0.871 & 0.454 & 0.876 & 0.324 & 0.434 \\
            
            MC-DropConnect & 0.3 & 0.859 & 0.822 & 0.218 & 0.338 & 0.407 & 0.336 & 0.337 & 0.451 & 0.000 & 0.961 & 0.518 & 0.895 & 0.002 & 0.918 & 0.649 & 0.876 & 0.445 & 0.841 & 0.298 & 0.462 \\
            
            MC-DropConnect & 0.4 & 0.865 & 0.843 & 0.201 & 0.435 & 0.230 & 0.491 & 
            0.397 & 0.505 & 0.000 & 0.972 & 0.576 & 0.863 & 0.003 & 0.914 & 0.637 & 0.898 & 0.460 & 0.827 & 0.302 & 0.331\\
            \bottomrule		
    \end{tabular}}
    \caption{Class-wise accuracy on 08 sequence of SemanticKITTI. The None value indicates the baseline model without any uncertainty quantification.}
    \label{table:darknet_perf_acc}
\end{table*}

\FloatBarrier
\section{Sample Point Cloud Visualizations}

\begin{figure}[!t]
    \centering
    \begin{tikzpicture}
        \node[] at (-13.8, 1.2) {Segmentation Class Labels};
        \begin{customlegend}[legend columns = 8,legend style = {column sep=1ex}, legend cell align = left,
        legend entries={Unlabeled, Car, Bicycle, Motorcycle, Truck, Other Vehicle, Person, Bicyclist, Motorcyclist, Road, Parking,
                        Sidewalk, Other Ground, Builing, Fence, Vegetation, Trunk, Terrain, Pole, Traffic Sign}]
        \addlegendimage{mark=none,unlabeled, only marks}
        \addlegendimage{mark=none,car, only marks}
        \addlegendimage{mark=none,bicycle, only marks}
        \addlegendimage{mark=none,motorcycle, only marks}
        \addlegendimage{mark=none,truck, only marks}
        \addlegendimage{mark=none,othervehicle, only marks}
        \addlegendimage{mark=none,person, only marks}
        \addlegendimage{mark=none,bicyclist, only marks}
        \addlegendimage{mark=none,motorcyclist, only marks}
        \addlegendimage{mark=none,road, only marks}
        \addlegendimage{mark=none,parking, only marks}
        \addlegendimage{mark=none,sidewalk, only marks}
        \addlegendimage{mark=none,otherground, only marks}
        \addlegendimage{mark=none,building, only marks}
        \addlegendimage{mark=none,fence, only marks}
        \addlegendimage{mark=none,vegetation, only marks}
        \addlegendimage{mark=none,trunk, only marks}
        \addlegendimage{mark=none,terrain, only marks}
        \addlegendimage{mark=none,pole, only marks}
        \addlegendimage{mark=none,trafficsign, only marks}        
        \end{customlegend}
    \end{tikzpicture}

    \begin{tikzpicture}
    \node[] at (-8.5, 1.2) {Entropy Values};
    \begin{customlegend}[legend columns = 5, legend style = {column sep=1ex}, legend cell align = left,
        legend entries={0 - 0.28, 0.29  - 0.56, 0.57 - 0.84, 0.85 - 1.12, 1.13 - 1.42, 1.43 - 1.70, 1.71 - 1.98, 1.99 - 2.26, 2.27 -  2.54, $>$ 2.54}]
        \addlegendimage{mark=none,entropyA, only marks}
        \addlegendimage{mark=none,entropyB, only marks}
        \addlegendimage{mark=none,entropyC, only marks}
        \addlegendimage{mark=none,entropyD, only marks}
        \addlegendimage{mark=none,entropyE, only marks}
        \addlegendimage{mark=none,entropyF, only marks}
        \addlegendimage{mark=none,entropyG, only marks}
        \addlegendimage{mark=none,entropyH, only marks}
        \addlegendimage{mark=none,entropyI, only marks}
        \addlegendimage{mark=none,entropyJ, only marks}
    \end{customlegend}
    \end{tikzpicture}
    
    \begin{subfigure}{0.47\textwidth}
        \centering
        \includegraphics[width=\textwidth]{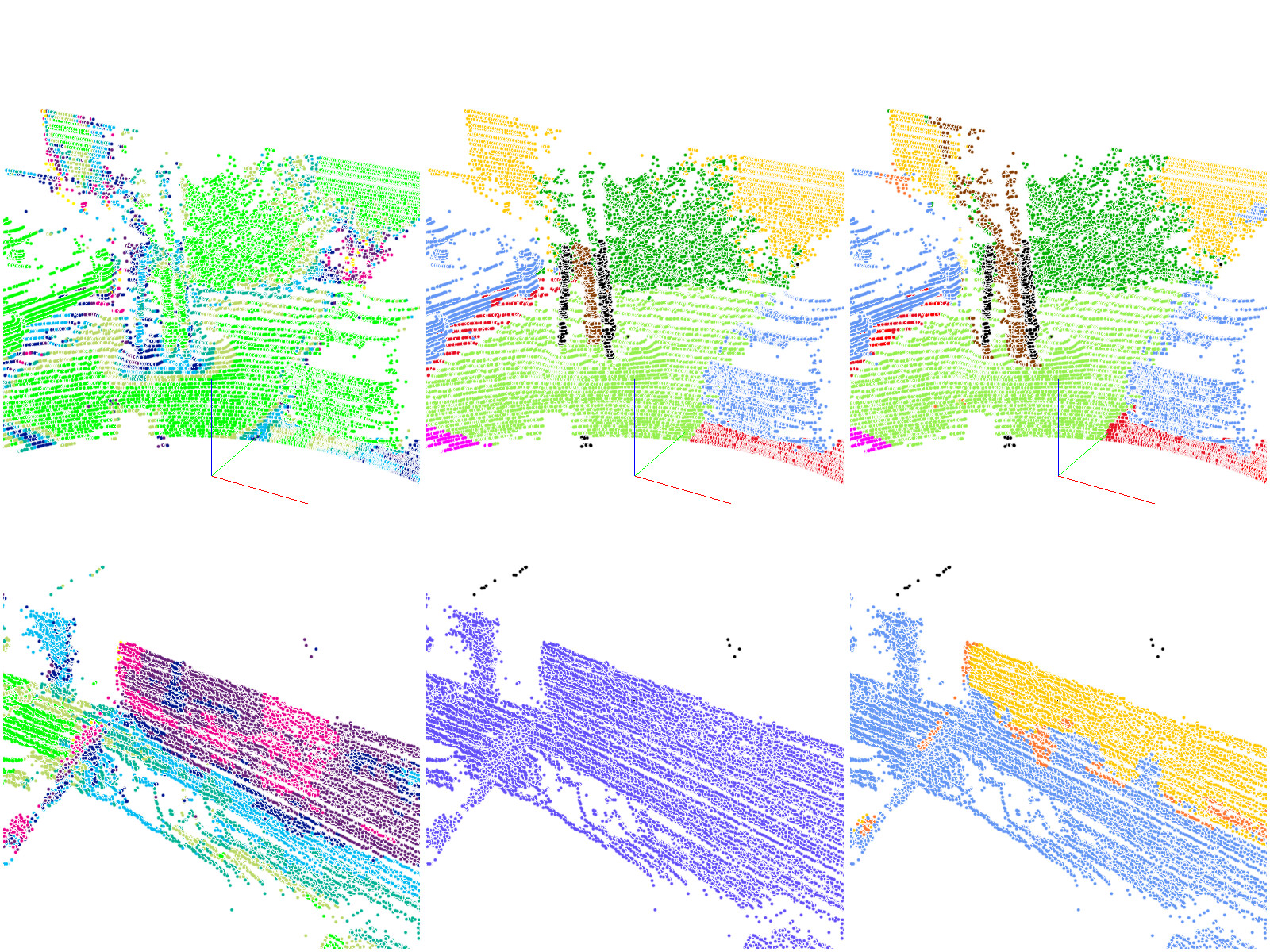}
        \caption{Point Cloud}
    \end{subfigure}
    \hspace*{0.1cm}
    \begin{subfigure}{0.47\textwidth}
        \centering
        \includegraphics[width=\textwidth]{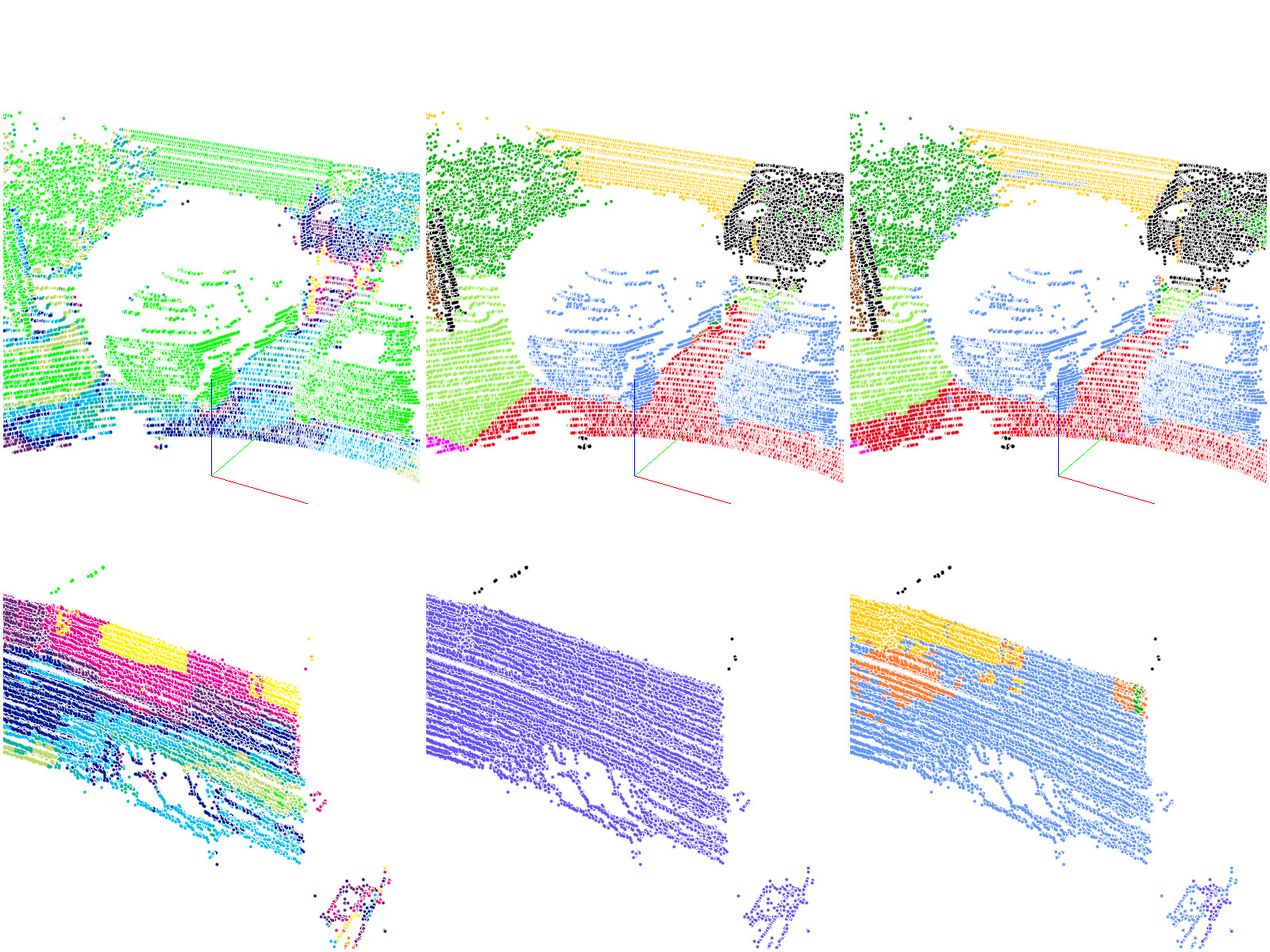}
        \caption{Point Cloud}
    \end{subfigure}
    \begin{subfigure}{0.47\textwidth}
        \centering
        \includegraphics[width=\textwidth]{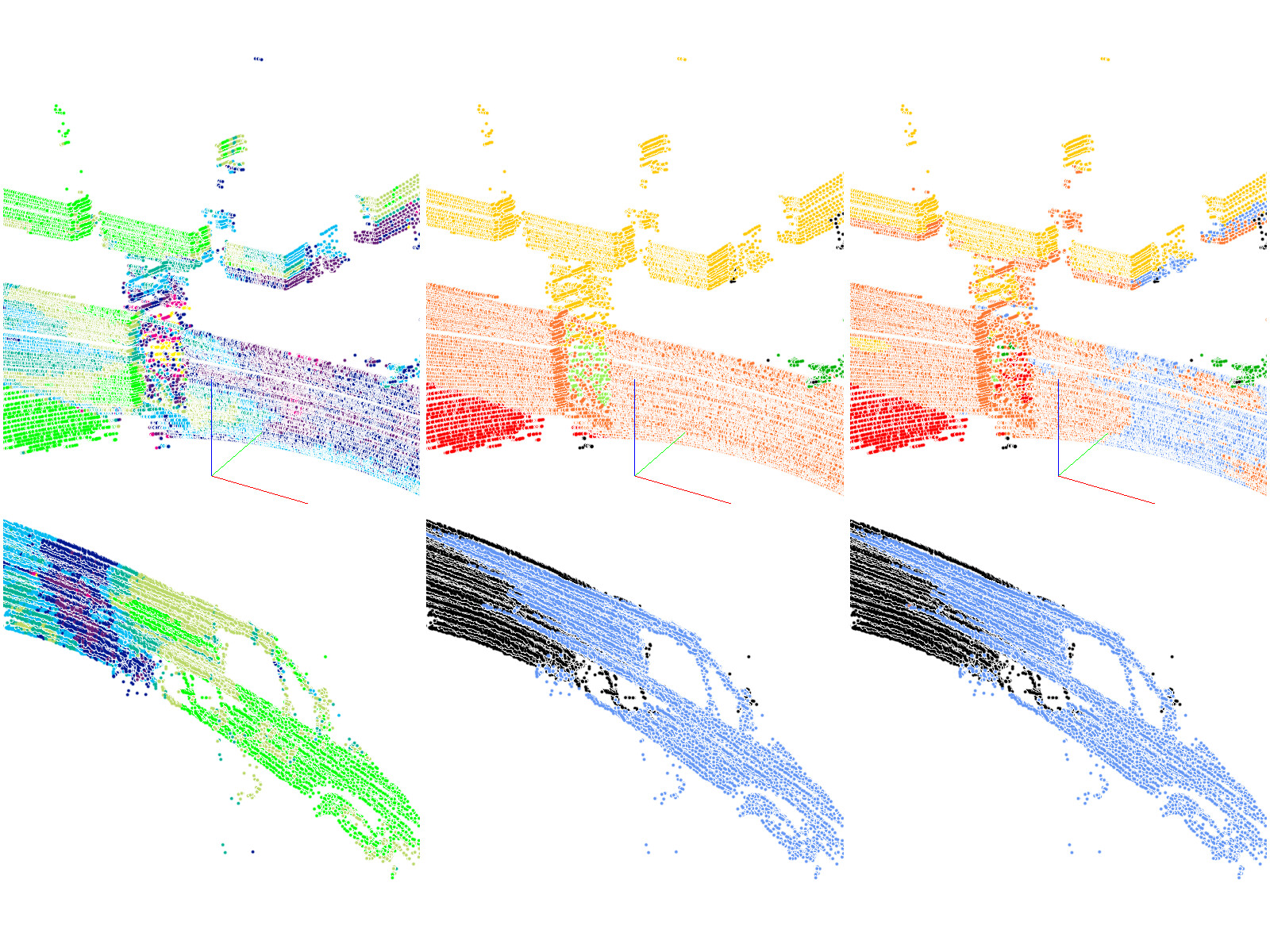}
        \caption{Point Cloud}
    \end{subfigure}
    \caption{Deep Ensembles - Per-point entropy visualization (left) compared to ground truth (center) and predicted segmentation (right)}
    \label{figure:vis_de}
\end{figure}
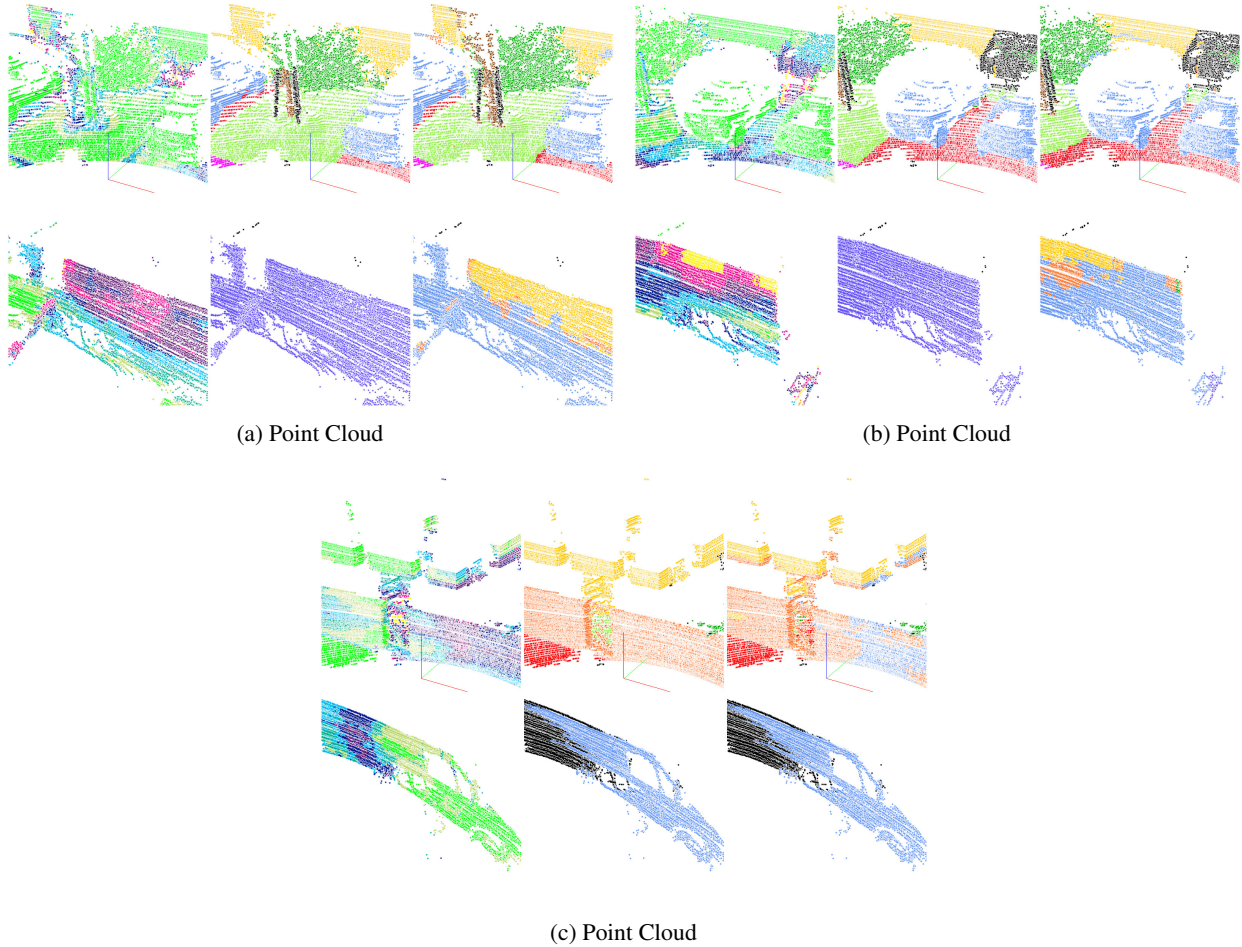

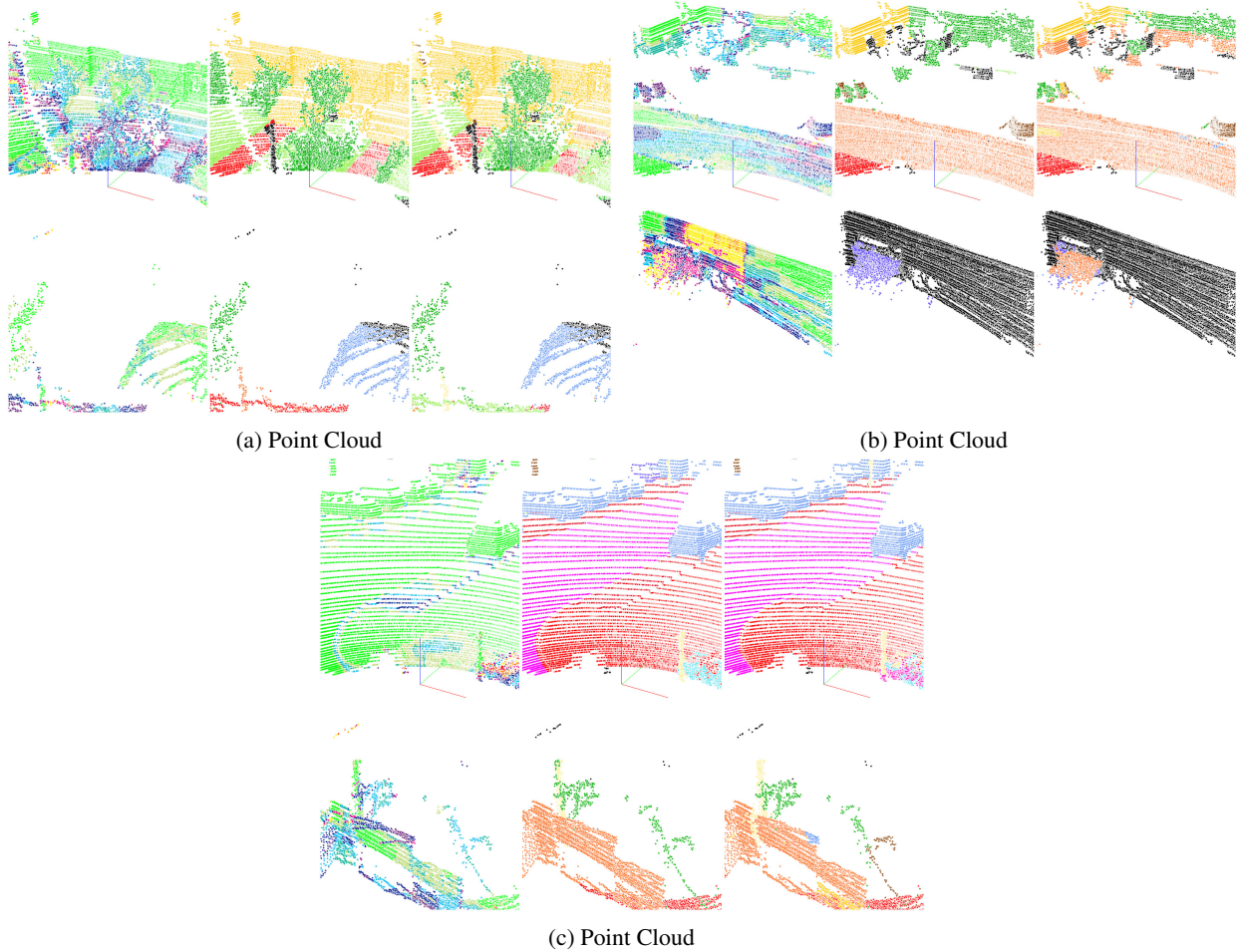
\begin{figure}[!t]
    \centering
    
    \begin{tikzpicture}
    \node[] at (-13.8, 1.2) {Segmentation Class Labels};
    \begin{customlegend}[legend columns = 8,legend style = {column sep=1ex}, legend cell align = left,
    legend entries={Unlabeled, Car, Bicycle, Motorcycle, Truck, Other Vehicle, Person, Bicyclist, Motorcyclist, Road, Parking,
        Sidewalk, Other Ground, Builing, Fence, Vegetation, Trunk, Terrain, Pole, Traffic Sign}]
    \addlegendimage{mark=none,unlabeled, only marks}
    \addlegendimage{mark=none,car, only marks}
    \addlegendimage{mark=none,bicycle, only marks}
    \addlegendimage{mark=none,motorcycle, only marks}
    \addlegendimage{mark=none,truck, only marks}
    \addlegendimage{mark=none,othervehicle, only marks}
    \addlegendimage{mark=none,person, only marks}
    \addlegendimage{mark=none,bicyclist, only marks}
    \addlegendimage{mark=none,motorcyclist, only marks}
    \addlegendimage{mark=none,road, only marks}
    \addlegendimage{mark=none,parking, only marks}
    \addlegendimage{mark=none,sidewalk, only marks}
    \addlegendimage{mark=none,otherground, only marks}
    \addlegendimage{mark=none,building, only marks}
    \addlegendimage{mark=none,fence, only marks}
    \addlegendimage{mark=none,vegetation, only marks}
    \addlegendimage{mark=none,trunk, only marks}
    \addlegendimage{mark=none,terrain, only marks}
    \addlegendimage{mark=none,pole, only marks}
    \addlegendimage{mark=none,trafficsign, only marks}        
    \end{customlegend}
    \end{tikzpicture}

    \begin{tikzpicture}
    \node[] at (-8.5, 1.2) {Entropy Values};
    \begin{customlegend}[legend columns = 5, legend style = {column sep=1ex}, legend cell align = left,
    legend entries={0 - 0.28, 0.29  - 0.56, 0.57 - 0.84, 0.85 - 1.12, 1.13 - 1.42, 1.43 - 1.70, 1.71 - 1.98, 1.99 - 2.26, 2.27 -  2.54, $>$ 2.54}]
    \addlegendimage{mark=none,entropyA, only marks}
    \addlegendimage{mark=none,entropyB, only marks}
    \addlegendimage{mark=none,entropyC, only marks}
    \addlegendimage{mark=none,entropyD, only marks}
    \addlegendimage{mark=none,entropyE, only marks}
    \addlegendimage{mark=none,entropyF, only marks}
    \addlegendimage{mark=none,entropyG, only marks}
    \addlegendimage{mark=none,entropyH, only marks}
    \addlegendimage{mark=none,entropyI, only marks}
    \addlegendimage{mark=none,entropyJ, only marks}
    \end{customlegend}
    \end{tikzpicture}
    \begin{subfigure}{0.47\textwidth}
        \centering
        \includegraphics[width=\textwidth]{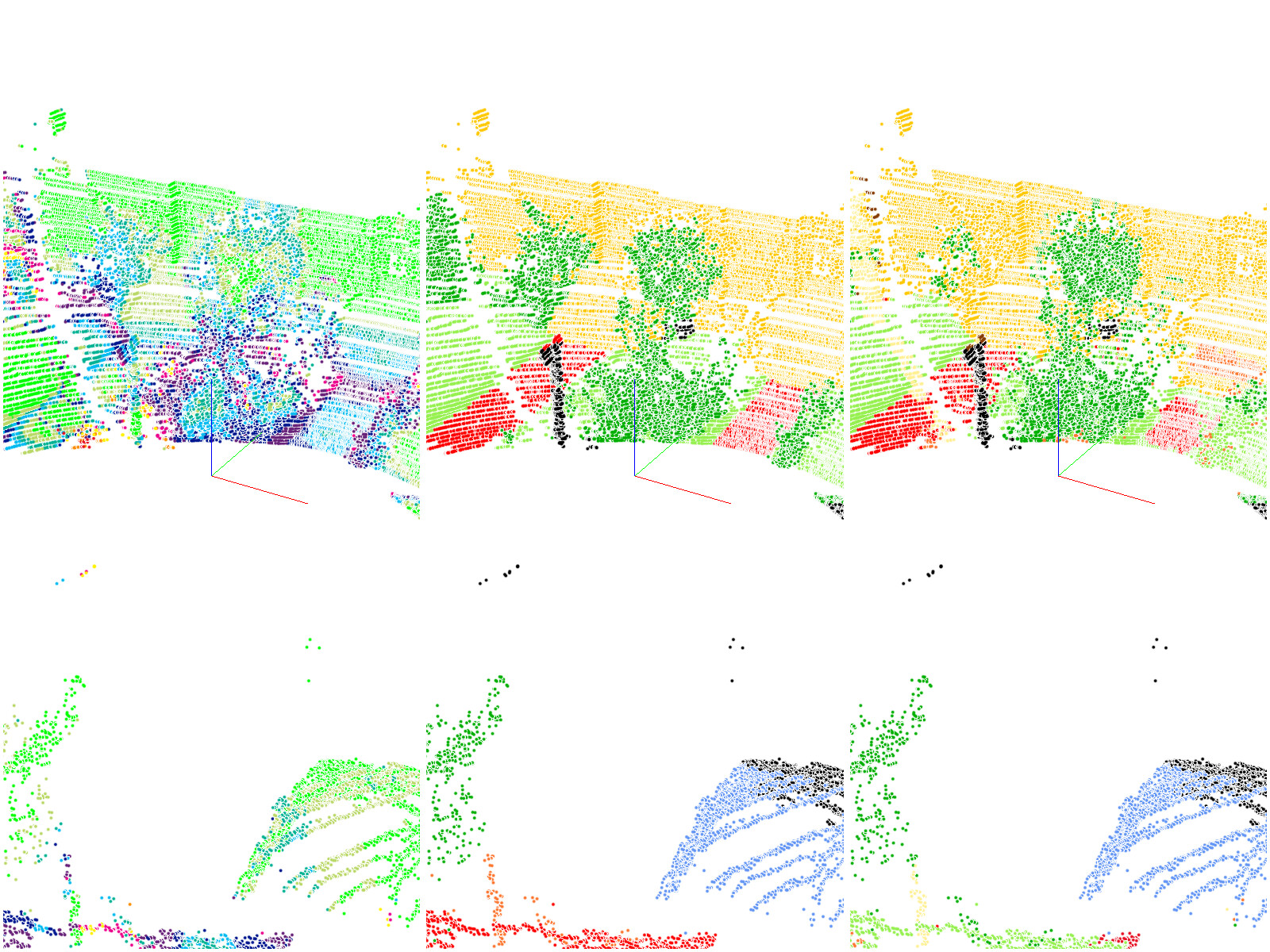}
        \caption{Point Cloud}
    \end{subfigure}
    \hspace*{0.1cm}
    \begin{subfigure}{0.47\textwidth}
        \centering
        \includegraphics[width=\textwidth]{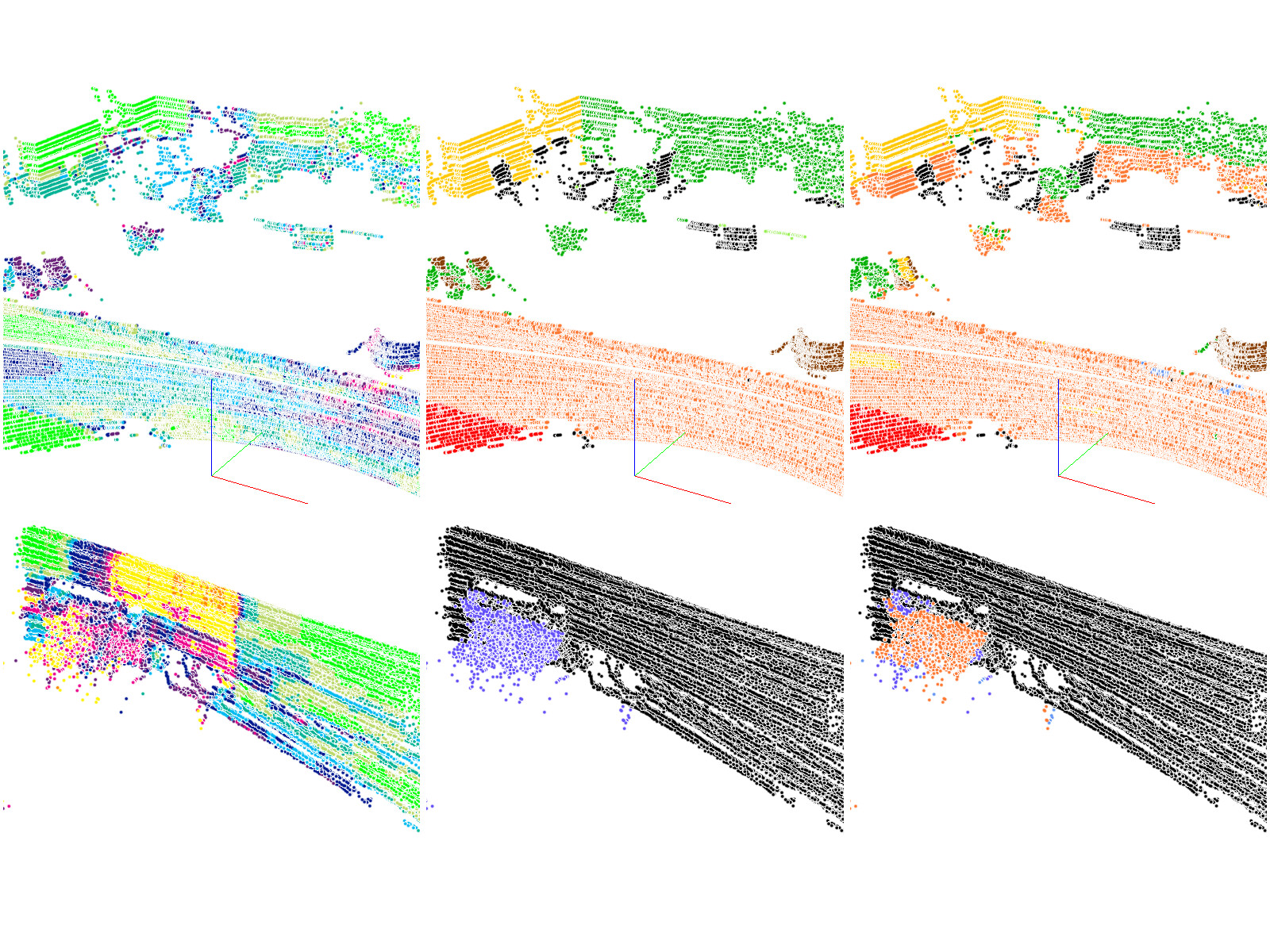}
        \caption{Point Cloud}
        \label{figure:vis_do_unlabeled}
    \end{subfigure}
    \begin{subfigure}{0.47\textwidth}
        \centering
        \includegraphics[width=\textwidth]{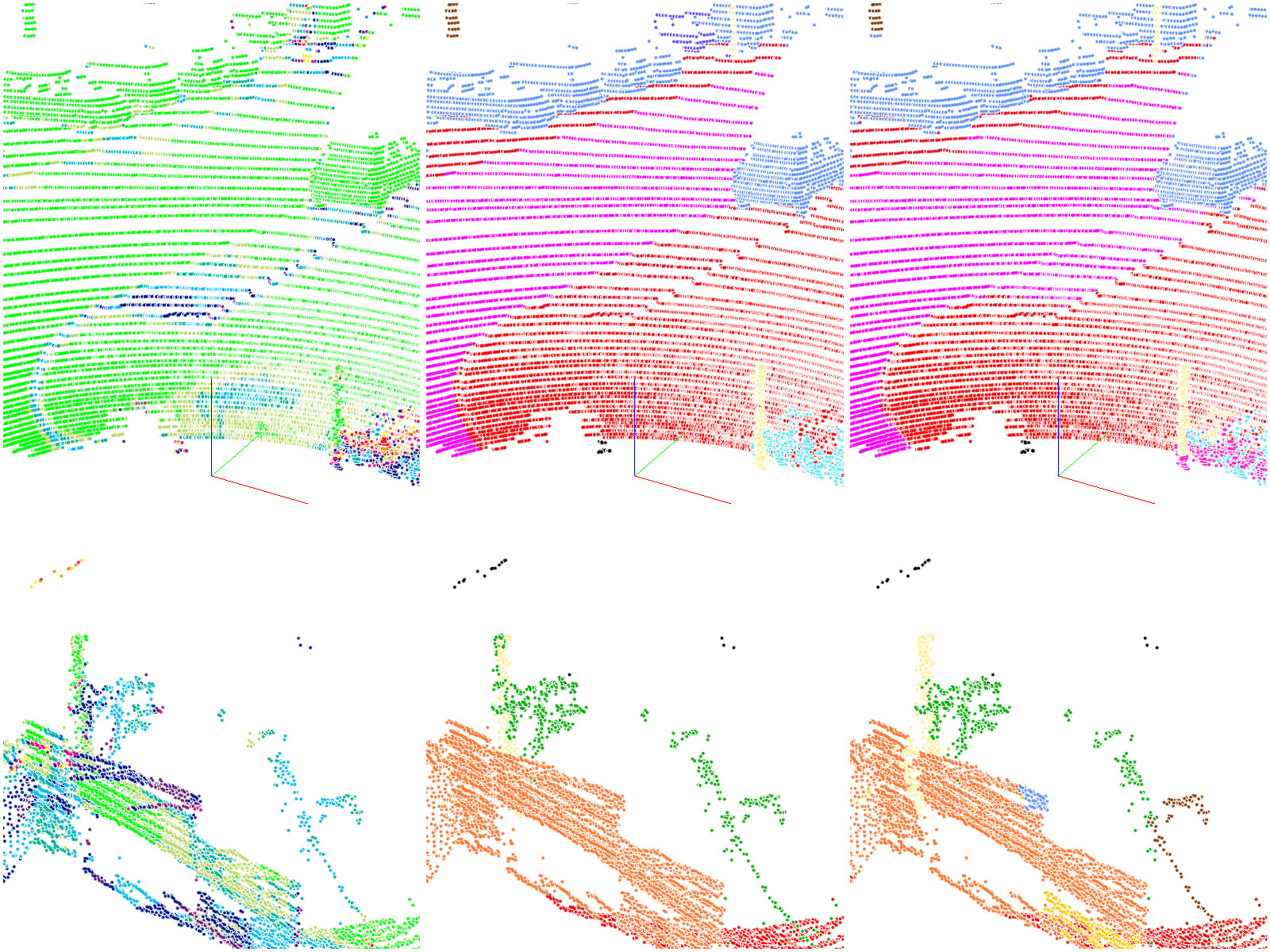}
        \caption{Point Cloud}
    \end{subfigure}
    \caption{MC Dropout - Per-point entropy visualization (left) compared to ground truth (center) and predicted segmentation (right)}
    \label{figure:vis_do}
\end{figure}

\begin{figure}[!t]
    \centering
    
    \begin{tikzpicture}
    \node[] at (-13.8, 1.2) {Segmentation Class Labels};
    \begin{customlegend}[legend columns = 8,legend style = {column sep=1ex}, legend cell align = left,
    legend entries={Unlabeled, Car, Bicycle, Motorcycle, Truck, Other Vehicle, Person, Bicyclist, Motorcyclist, Road, Parking, Sidewalk, Other Ground, Builing, Fence, Vegetation, Trunk, Terrain, Pole, Traffic Sign}]
    \addlegendimage{mark=none,unlabeled, only marks}
    \addlegendimage{mark=none,car, only marks}
    \addlegendimage{mark=none,bicycle, only marks}
    \addlegendimage{mark=none,motorcycle, only marks}
    \addlegendimage{mark=none,truck, only marks}
    \addlegendimage{mark=none,othervehicle, only marks}
    \addlegendimage{mark=none,person, only marks}
    \addlegendimage{mark=none,bicyclist, only marks}
    \addlegendimage{mark=none,motorcyclist, only marks}
    \addlegendimage{mark=none,road, only marks}
    \addlegendimage{mark=none,parking, only marks}
    \addlegendimage{mark=none,sidewalk, only marks}
    \addlegendimage{mark=none,otherground, only marks}
    \addlegendimage{mark=none,building, only marks}
    \addlegendimage{mark=none,fence, only marks}
    \addlegendimage{mark=none,vegetation, only marks}
    \addlegendimage{mark=none,trunk, only marks}
    \addlegendimage{mark=none,terrain, only marks}
    \addlegendimage{mark=none,pole, only marks}
    \addlegendimage{mark=none,trafficsign, only marks}        
    \end{customlegend}
    \end{tikzpicture}

    \begin{tikzpicture}
    \node[] at (-8.5, 1.2) {Entropy Values};
    \begin{customlegend}[legend columns = 5, legend style = {column sep=1ex}, legend cell align = left,
    legend entries={0 - 0.28, 0.29  - 0.56, 0.57 - 0.84, 0.85 - 1.12, 1.13 - 1.42, 1.43 - 1.70, 1.71 - 1.98, 1.99 - 2.26, 2.27 -  2.54, $>$ 2.54}]
    \addlegendimage{mark=none,entropyA, only marks}
    \addlegendimage{mark=none,entropyB, only marks}
    \addlegendimage{mark=none,entropyC, only marks}
    \addlegendimage{mark=none,entropyD, only marks}
    \addlegendimage{mark=none,entropyE, only marks}
    \addlegendimage{mark=none,entropyF, only marks}
    \addlegendimage{mark=none,entropyG, only marks}
    \addlegendimage{mark=none,entropyH, only marks}
    \addlegendimage{mark=none,entropyI, only marks}
    \addlegendimage{mark=none,entropyJ, only marks}
    \end{customlegend}
    \end{tikzpicture}
    
    \begin{subfigure}{0.47\textwidth}
        \centering
        \includegraphics[width=\textwidth]{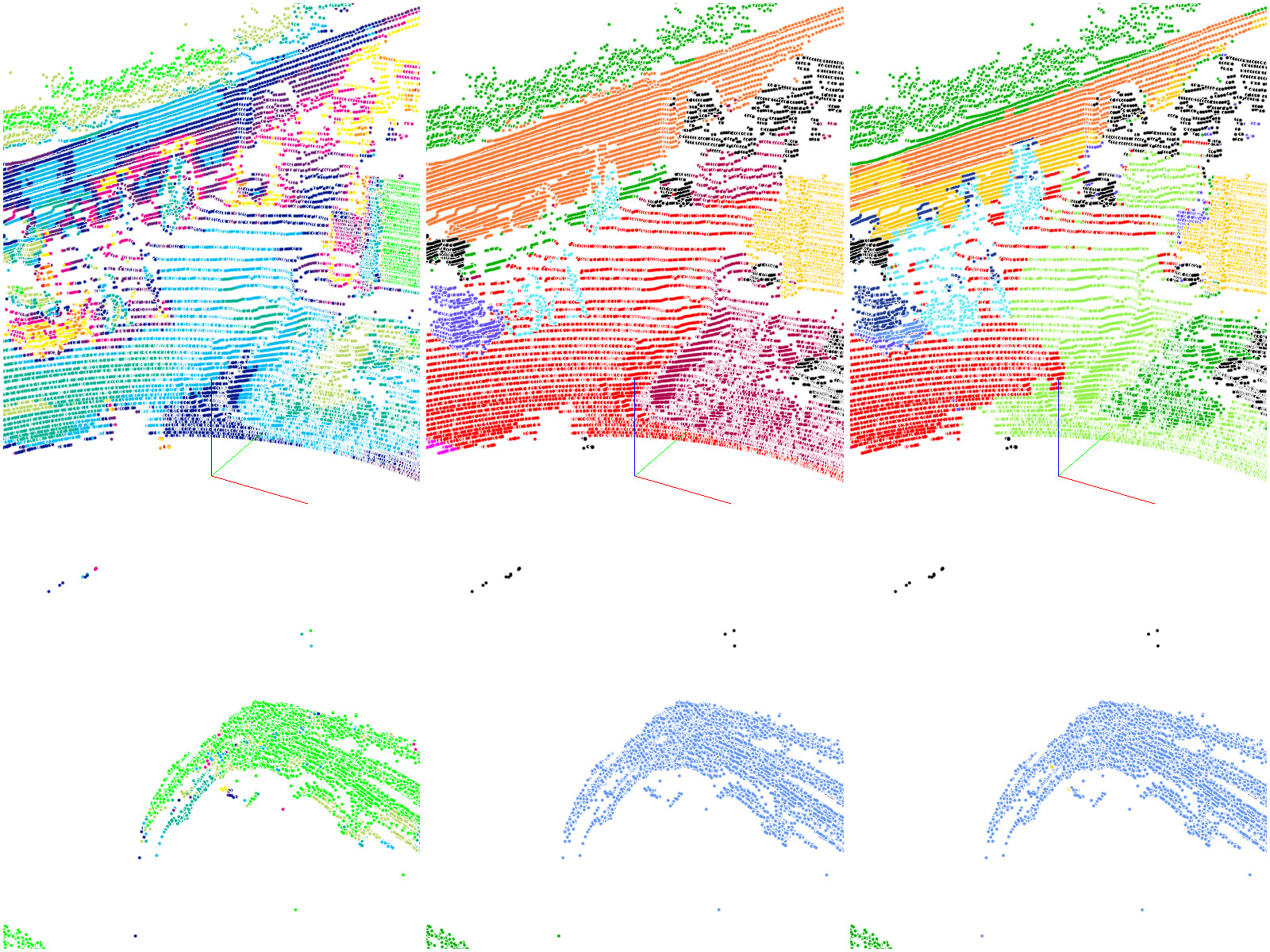}
        \caption{Point Cloud}
    \end{subfigure}
    \hspace*{0.1cm}
    \begin{subfigure}{0.47\textwidth}
        \centering
        \includegraphics[width=\textwidth]{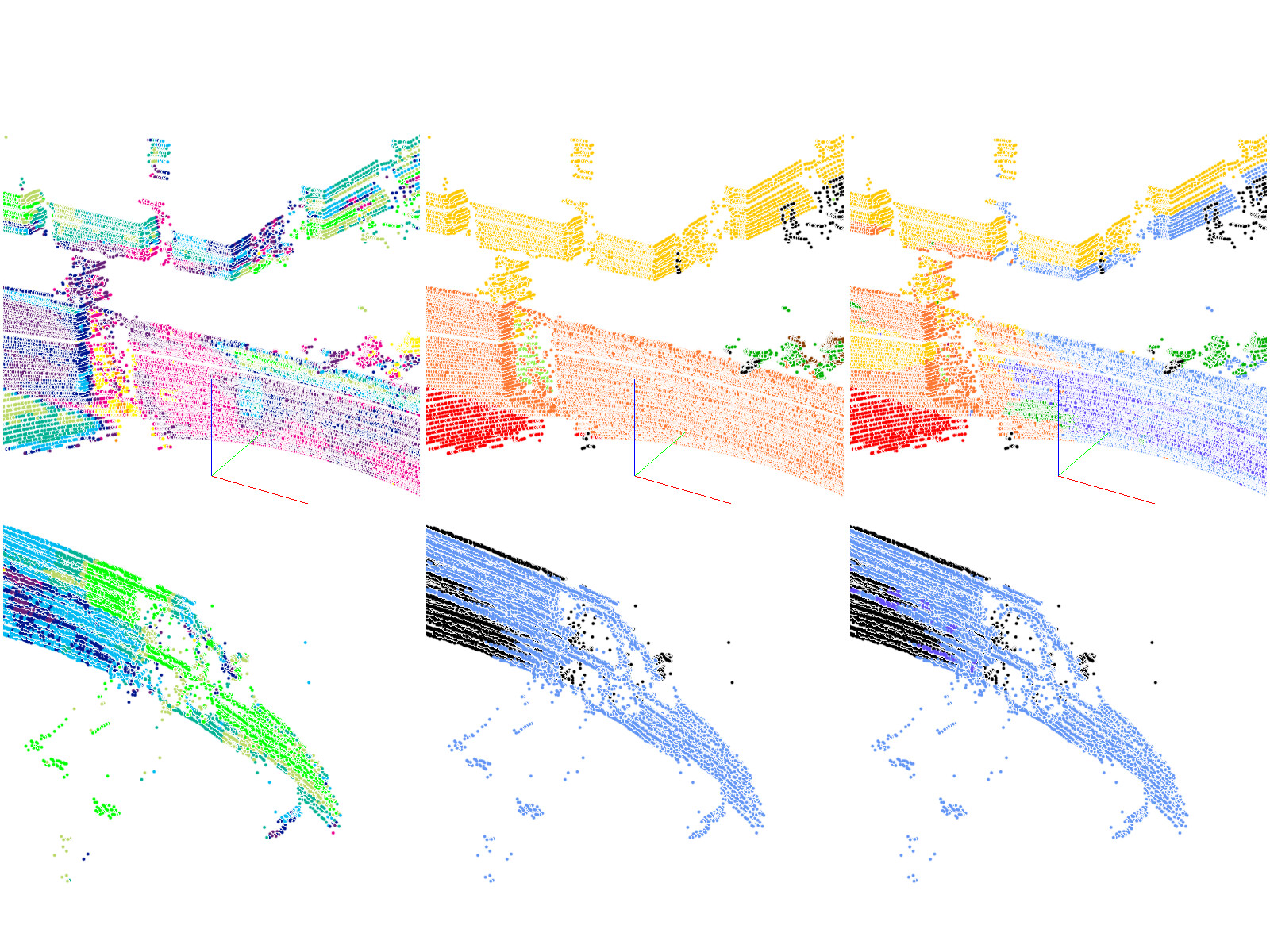}
        \caption{Point Cloud}
        \label{figure:vis_dc_unlabeled}
    \end{subfigure}
    \begin{subfigure}{0.47\textwidth}
        \centering
        \includegraphics[width=\textwidth]{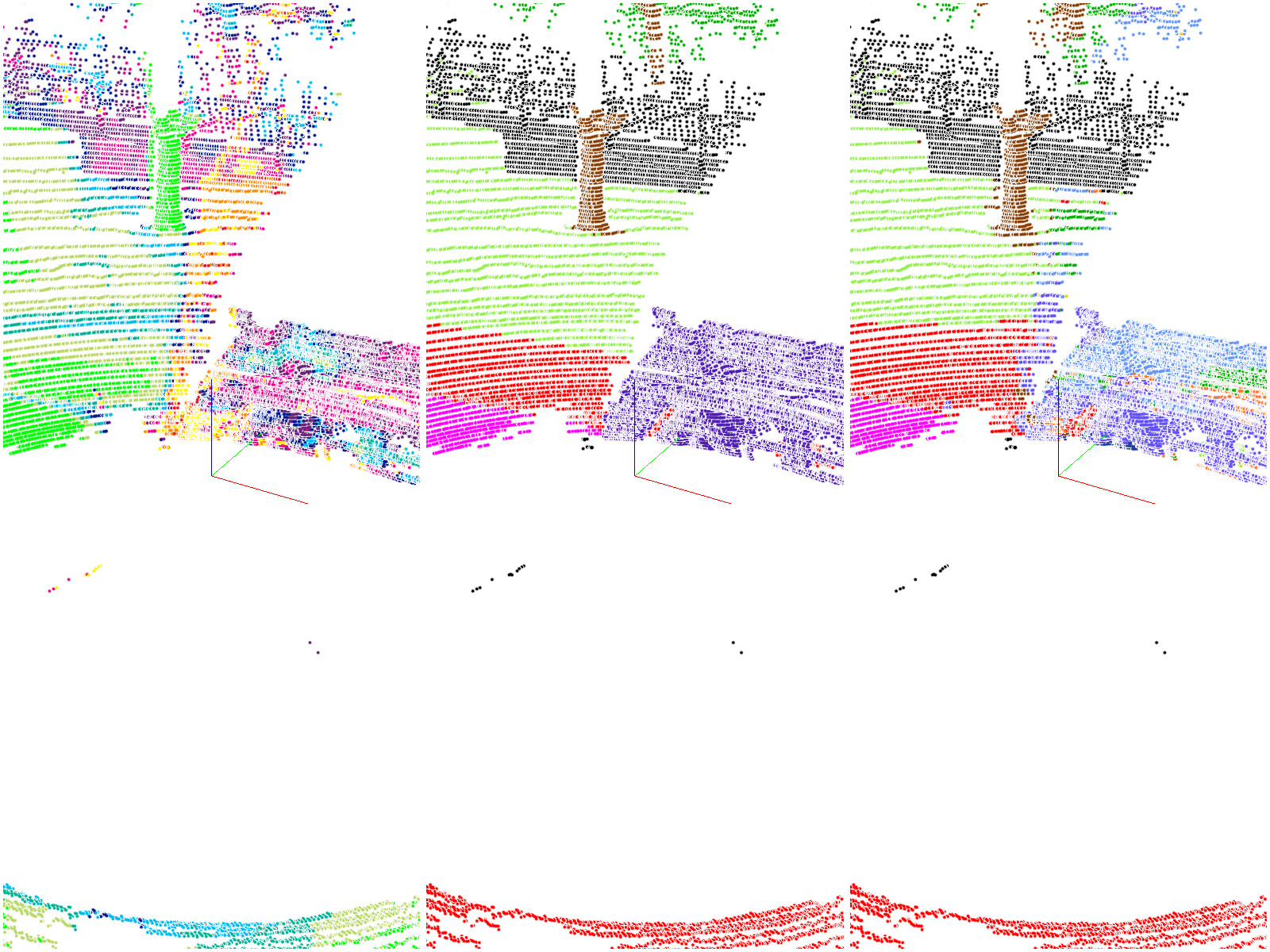}
        \caption{Point Cloud}
    \end{subfigure}
    \caption{MC DropConnect - Per-point entropy visualization (left) compared to ground truth (center) and predicted segmentation (right)}
    \label{figure:vis_dc}
\end{figure}
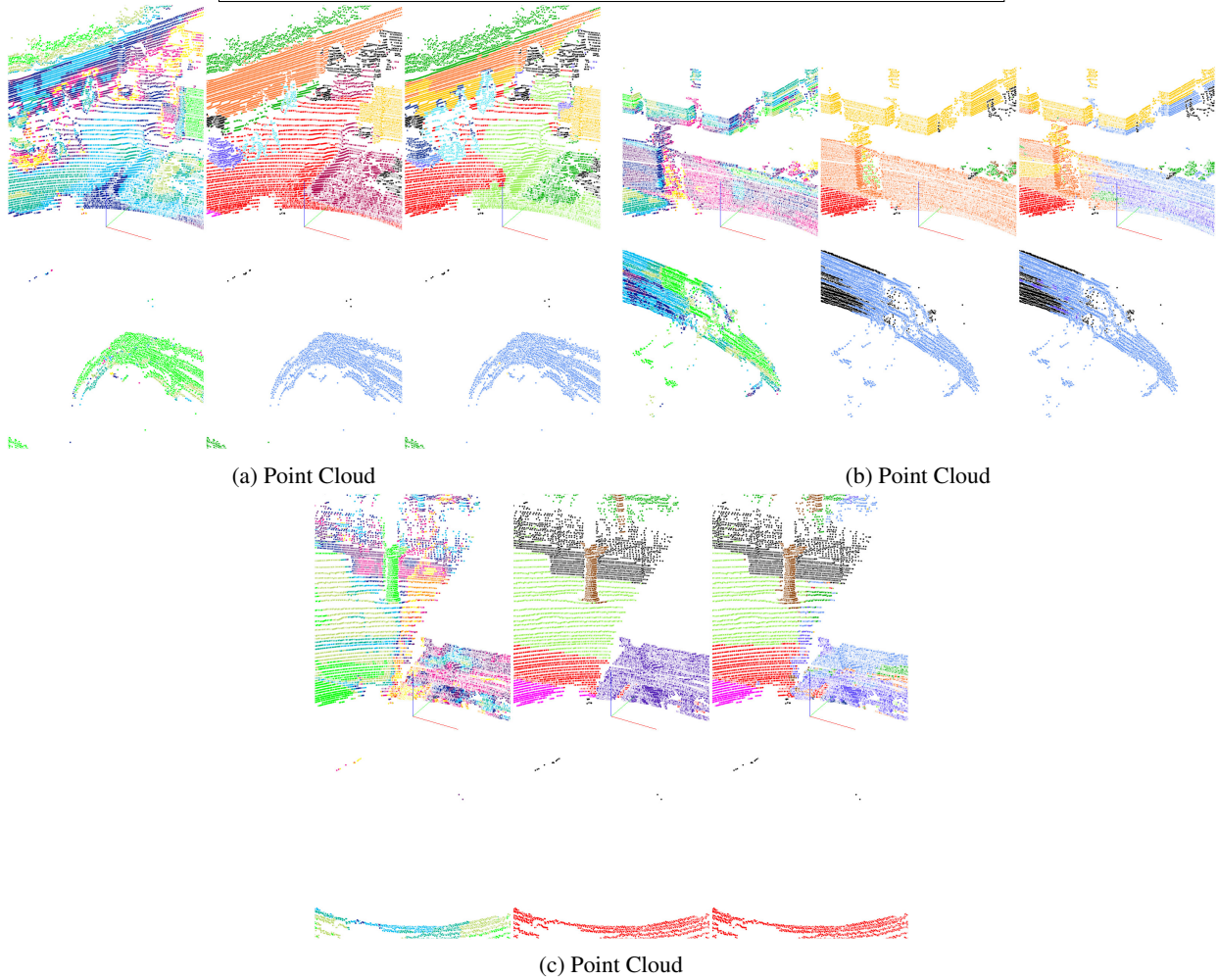

In this section we provide visualizations of predicted segmentations and their uncertainty for each method. We selected point clouds according to the entropy predicted by each method. The top three point cloud scans are shown in Figure \ref{figure:vis_de}, Figure \ref{figure:vis_do}, and Figure \ref{figure:vis_dc}.

Overall all three methods produce increasing uncertainty for points in between class regions. This is particularly stronger for some classes such as sidewalk vs car, vegetation vs building/terrain, terrain vs sidewalk. The classes representing "other" kinds of objects (such as other vehicle and other ground) generally have higher uncertainty and incorrect segmentations, which can be expected due to their large variability.

\newpage
The unlabeled class has large sections of high uncertainty, for example as it can be seen in Figure \ref{figure:vis_do_unlabeled} and Figure \ref{figure:vis_dc_unlabeled}. Specific to points clouds, we see a pattern that regions with less points are more uncertain, which indicates that the model considers local information in the point cloud into building their uncertainty estimate. It makes sense that the lack of information (as presented with missing points in the cloud) would produce higher uncertainty, which also correlates with some classes that normally have small number of points (such as bicyclist, motorcycle, and bicycle) due to the small object size.

\section{Acknowledgments}

We gratefully acknowledge the continued support by the b-it Bonn-Aachen International Center for Information Technology
and the Hochschule Bonn-Rhein-Sieg.

\end{document}